\documentclass[10pt]{article} 
\usepackage[accepted]{tmlr}

\usepackage{amsmath,amsfonts,bm}

\usepackage{amssymb,mathtools}
\usepackage{bbm}
\newcommand{\indicator}[1]{\mathbf{1}_{#1}}
\newcommand{\balpha}{\bm{\alpha}}
\newcommand{\bbeta}{\bm{\beta}}
\newcommand{\btheta}{\bm{\theta}}

\newcommand{\bphi}{\bm{\phi}}

\newcommand{\bx}{\mathbf{x}}
\newcommand{\by}{\mathbf{y}}
\newcommand{\bz}{\mathbf{z}}
\newcommand{\bchi}{\bm{\chi}}
\newcommand{\bpi}{\bm{\pi}}
\newcommand{\bpsi}{\bm{\psi}}

\def\eqref#1{equation~\ref{#1}}

\def\1{\bm{1}}

\DeclareMathAlphabet{\mathsfit}{\encodingdefault}{\sfdefault}{m}{sl}
\SetMathAlphabet{\mathsfit}{bold}{\encodingdefault}{\sfdefault}{bx}{n}

\DeclareMathOperator*{\argmax}{arg\,max}

\usepackage[hidelinks]{hyperref}
\usepackage{url}

\usepackage{cancel}
\usepackage{pifont}
\usepackage{tikz}
\newcommand{\cmark}{\ding{51}}%
\newcommand{\xmark}{\ding{55}}%
\usepackage{booktabs}

\DeclarePairedDelimiterXPP\Expect[2]{\mathbb{E}_{#1}}[]{}{#2}%

\newcommand{\alignedintertext}[1]{%
  \noalign{%
    \vskip\belowdisplayshortskip
    \vtop{\hsize=\linewidth#1\par
    \expandafter}%
    \expandafter\prevdepth\the\prevdepth
  }%
}

\newcommand*\circled[1]{\tikz[baseline=(char.base)]{
            \node[shape=circle,draw,inner sep=.6pt] (char) {#1};}}

\makeatletter
\newcommand{\vast}{\bBigg@{4}}
\newcommand{\Vast}{\bBigg@{5}}
\makeatother

\usepackage{url}

\usepackage{breakurl}

\allowdisplaybreaks

\usepackage{makecell}
\usepackage{wrapfig}
\usepackage{subcaption}
\usepackage{comment}
\usepackage[noabbrev,capitalize,nameinlink]{cleveref}

\usepackage{pdflscape}

\usepackage{fontawesome5}

\title{Prior and Posterior Networks: A Survey on Evidential Deep Learning Methods For Uncertainty Estimation}

\author{\name Dennis Ulmer\textsuperscript{\faCompass} \email dennis.ulmer@mailbox.org \\
       \name Christian Hardmeier\textsuperscript{\faCompass} \email chrha@itu.dk  \\
       \name Jes Frellsen\textsuperscript{\faRobot,\faCompressArrows*} \email jefr@dtu.dk \\
       \AND
       \addr \textsuperscript{\faCompass}IT University of Copenhagen, \textsuperscript{\faRobot}Technical University of Denmark, \textsuperscript{\faCompressArrows*}Pioneer Centre for Artificial Intelligence}

\begin{document}

\maketitle

\begin{abstract}
Popular approaches for quantifying predictive uncertainty in deep neural networks often involve distributions over weights or multiple models, for instance via Markov Chain sampling,  ensembling, or Monte Carlo dropout. These techniques usually incur overhead by having to train multiple model instances or do not produce very diverse predictions. This comprehensive and extensive survey aims to familiarize the reader with an alternative class of models based on the concept of \emph{Evidential Deep Learning}: For unfamiliar data, they aim to admit ``what they don't know'', and fall back onto a prior belief. Furthermore, they allow uncertainty estimation in a single model and forward pass by parameterizing \emph{distributions over distributions}. This survey recapitulates existing works, focusing on the implementation in a classification setting, before surveying the application of the same paradigm to regression. We also reflect on the strengths and weaknesses compared to other existing methods and provide the most fundamental derivations using a unified notation to aid future research.
\end{abstract}

\section{Introduction}

\begin{wrapfigure}[20]{r}{0.5\textwidth}
    \centering
    \vspace{-1.3cm}
    \includegraphics[width=0.48\textwidth]{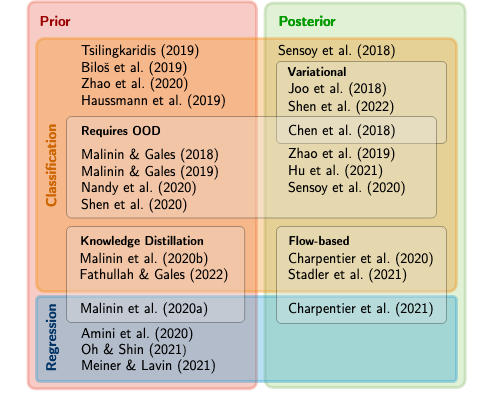}
     \caption{Taxonomy of surveyed approaches, divided into tractable parameterizations of the prior or posterior on one axis (see \cref{tab:overview-prior,tab:overview-posterior} for an overview) and into approaches for classification and regression on the other. Regression methods are outlined in \cref{table:overview-regression}.}\label{fig:taxonomy}
\end{wrapfigure}

Many existing methods for uncertainty estimation leverage the concept of Bayesian model averaging: 
These include ensembling \citep{lakshminarayanan2017simple, wilson2020bayesian}, Markov chain Monte Carlo sampling \citep{de2003bayesian, andrieu2013reversible} as well as variational inference approaches \citep{mackay1992bayesian, mackay1995developments, hinton1993keeping, neal2012bayesian}, 
including approaches such as Monte Carlo (MC) dropout \citep{gal2016dropout} and Bayes-by-backprop \citep{blundell2015weight}. Bayesian model averaging for neural networks usually involves the approximation of an otherwise infeasible integral using MC samples. This causes the following problems: Firstly, the quality of the MC approximation depends on the veracity and diversity of samples from the weight posterior. Secondly, the approach often involves increasing the number of parameters in a model or training more model instances altogether. Recently, a new class of models has been proposed to side-step this conundrum by using a different factorization of the posterior predictive distribution. This allows computing uncertainty in a single forward pass and with a single set of weights. These models are grounded in a concept coined \emph{Evidential Deep Learning}: For out-of-distribution (OOD) inputs, they are encouraged to fall back onto a prior. This is often described as \emph{knowing what they don't know}.

In this paper, we summarize the existing literature and provide an overview of Evidential Deep Learning approaches. We give an overview over all discussed work in  \cref{fig:taxonomy}, where we distinguish surveyed works for classification between models parameterizing a Dirichlet prior (\cref{sec:prior-networks}) or posterior (\cref{sec:posterior-networks}). We further discuss similar methods for regression problems (\cref{sec:evidential-regression}). As we will see, obtaining well-behaving uncertainty estimates can be challenging in the Evidential Deep Learning framework; proposed solutions that are also reflected in \cref{fig:taxonomy} are the usage of OOD examples during training \citep{malinin2018predictive, malinin2019reverse, nandy2020towards, shen2020modeling, chen2018variational, zhao2019quantifying, hu2021multidimensional, sensoy2020uncertainty}, knowledge distillation \citep{malinin2020ensemble, malinin2020regression} or the incorporation of density estimation \citep{charpentier2020posterior,charpentier2021natural,stadler2021graph}, which we discuss in more detail in \cref{sec:discussion}. This survey aims to both serve as an accessible introduction to this model family to the unfamiliar reader as well as an informative overview, in order to promote a wider application outside the uncertainty quantification literature. We also provide a collection of the most important derivations for the Dirichlet distribution for Machine Learning, which plays a central role in many of the discussed approaches. 

\section{Background}\label{sec:background}


We first introduce the central concepts for this survey, including Bayesian inference in \cref{sec:bayesian-inference}, Bayesian model averaging in \cref{sec:uncertainty-quantification} and Evidential Deep Learning in \cref{sec:evidential-deep-learning}.\footnote{Note that in the following we will use the suggested notation of the TMLR journal, e.g.\@ by using $P$ for probability mass and $p$ for probability density functions.}

\subsection{Bayesian Inference}\label{sec:bayesian-inference}

The foundation of the following sections is Bayesian inference: Given some prior belief $p(\btheta)$ about parameters of interest $\btheta$, we use available observations $\mathbb{D} = \{(x_i, y_i)\}_{i=1}^N$ and their likelihood $p(\mathbb{D}|\btheta)$ to obtain an updated belief in form of the posterior $p(\btheta|\mathbb{D}) \propto p(\mathbb{D}|\btheta)p(\btheta)$. This update rule is derived from Bayes' rule, namely

\begin{equation}\label{eq:bayes-rule}
    p(\btheta|\mathbb{D}) = \frac{p(\mathbb{D}|\btheta)p(\btheta)}{p(\mathbb{D})} = \frac{p(\mathbb{D}|\btheta)p(\btheta)}{\int p(\mathbb{D}|\btheta)p(\btheta) d\btheta},
\end{equation}

where we often try to avoid computing the term in the denominator since marginalization over a large (continuous) parameter space of $\btheta$ is usually intractable. In order to perform a prediction $y$ for a new data point $\bx$, we can now utilize the \emph{posterior predictive distribution} defined as 

\begin{equation}\label{eq:bma}
    P(y|\bx, \mathbb{D}) = \int P(y|\bx, \bm{\theta})p(\bm{\theta}|\mathbb{D})d\bm{\theta}.
\end{equation}

Since we integrate over the entire parameter space of $\btheta$, weighting each prediction by the posterior probability of its parameters to obtain the final result, this process is referred to as \emph{Bayesian model averaging} (BMA). Here,  predictions $P(y|\bx, \bm{\theta})$ stemming from parameters that are plausible given the observed data will receive a higher weight $p(\bm{\theta}|\mathbb{D})$ in the final prediction $P(y|\bx, \mathbb{D})$. As we will see in the following section, this factorization of the predictive predictive distribution also has beneficial properties for analyzing the uncertainty of a model. 

\subsection{Predictive Uncertainty in Neural Networks}\label{sec:uncertainty-quantification}

In probabilistic modelling, uncertainty is commonly divided into aleatoric and epistemic uncertainty \citep{der2009aleatory, kendall2017uncertainties, huellermeier2021aleatoric}. Aleatoric uncertainty refers to the uncertainty that is induced by the data-generating process, for instance noise or inherent overlap between observed instances of classes. Epistemic uncertainty is the type of uncertainty about the optimal model parameters (or even hypothesis class). It is reducible with an increasing amount of data, as fewer and fewer possible models become a plausible fit. These two notions resurface when formulating the posterior predictive distribution for a new data point $\bx$:\footnote{Note that the predictive distribution in  \cref{eq:bma} generalizes the common case for a single network prediction where $P(y|\bx, \mathbb{D}) \approx P(y|\bx, \hat{\bm{\theta}})$. Mathematically, this is expressed by replacing the posterior $p(\bm{\theta}|\mathbb{D})$ by a Dirac delta distribution as in \cref{eq:prior-networks-fac}, where all probability density rests on a single parameter configuration.}

\begin{equation}
    P(y|\bx, \mathbb{D}) = \int \underbrace{P(y|\bx, \bm{\theta})}_{\text{Aleatoric}}\underbrace{p(\bm{\theta}|\mathbb{D})}_{\text{Epistemic}}d\bm{\theta}.
\end{equation}

Here, the first factor captures the aleatoric uncertainty about the correct prediction, while the second one expresses uncertainty about the correct model parameters---the more data we observe, the more density of  $p(\bm{\theta}|\mathbb{D})$ should lie on reasonable parameter values for $\bm{\theta}$. For high-dimensional real-valued parameters $\bm{\theta}$ like in neural networks, this integral becomes intractable, and is usually approximated using Monte Carlo samples:\footnote{For easier distributions, the integral can often be evaluated analytically exploiting conjugacy. Another approach for more complex distributions can be the method of moments (see e.g. \citealp{duan2021method}).}

\begin{equation}
    P(y|\bx, \mathbb{D}) \approx \frac{1}{K}\sum_{k=1}^K P(y|\bx, \btheta^{(k)});\quad \btheta^{(k)} \sim p(\btheta|\mathbb{D})
\end{equation}

based on $K$ different sets of parameters $\btheta^{(k)}$. Since this requires obtaining multiple versions of model parameters through some additional procedure, this however comes with the aforementioned problems of computational overhead and approximation errors, motivating the approaches discussed in this survey. 

\subsection{Evidential Deep Learning}\label{sec:evidential-deep-learning}

Since the traditional approach to predictive uncertainty estimation requires multiple parameter sets and can only approximate the predictive posterior, we can factorize \cref{eq:bma} further to obtain a tractable form:

\begin{equation}\begin{aligned}\label{eq:prior-networks-fac}
    p(y|\bx, \mathbb{D}) & = \iint \underbrace{P(y|\bm{\pi})}_{\vphantom{\big[}\text{Aleatoric}}\underbrace{p(\bm{\pi}|\bx, \bm{\theta})}_{\vphantom{\big[}\text{\ Distributional\ }}\underbrace{p(\bm{\theta}|\mathbb{D})}_{\vphantom{\big[}\text{Epistemic}}d\bm{\pi} d\bm{\theta} \approx \int P(y|\bm{\pi})\underbrace{\vphantom{\big[}p(\bm{\pi}|\bx, \hat{\bm{\theta}})}_{ p(\bm{\theta}|\mathbb{D}) \approx \delta(\bm{\theta}-\hat{\bm{\theta}})}d\bm{\pi}.
\end{aligned}\end{equation}

This factorization contains another type of uncertainty, which \citet{malinin2018predictive} call the \emph{distributional} uncertainty, uncertainty caused by the mismatch of training and test data distributions. In the last step, \citet{malinin2018predictive} replace $p(\bm{\theta}|\mathbb{D})$ by a point estimate $\hat{\bm{\theta}}$ using the Dirac delta function, i.e.\@ a single trained neural network, to get rid of the intractable integral. Although another integral remains, retrieving the uncertainty from this predictive distribution actually has a closed-form analytical solution for the Dirichlet (see \cref{sec:uncertainty-dirichlet}). The advantage of this approach is further that it allows us to distinguish uncertainty about a data point because it is ambiguous from points coming from an entirely different data distribution. As an example, consider a binary classification problem, in which the data manifold consists of two overlapping clusters. As we are classifying a new data point, we obtain a distribution $P(y|\bx, \bm{\theta})$ which is uniform over the two classes. What does this mean? The model might either be confident that the point lies in the region of overlap and is inherently ambiguous, or that the model is uncertain about the correct class. Without further context, we cannot distinguish between these two cases \citep{bengs2022difficulty, hullermeier2022quantifying}. Compare that to instead predicting $p(\bm{\pi}|\bx, \bm{\theta})$: If the data point is ambiguous, the resulting distribution will be centered on $0.5$, if the model is generally uncertain, the distribution will be uniform, allowing this distinction. We will illustrate this principle further in the upcoming \cref{sec:example,sec:uncertainty-dirichlet}.

In the neural network context in \cref{eq:prior-networks-fac}, it should be noted that restricting oneself to a point estimate of the parameters prevent the estimation of epistemic uncertainty like in earlier works through the weight posterior $p(\bm{\theta}|\mathbb{D})$, as discussed in the next section. However, there are works like \citet{haussmann2019bayesian, zhao2020uncertainty} that combine both approaches.

The term \emph{Evidential Deep Learning} (EDL) originates from the work of \citet{sensoy2018evidential} and is based on the \emph{Theory of Evidence} \citep{dempster1968generalization,audun2018subjective}: Within the theory, belief mass is assigned to set of possible states, e.g.\@ class labels, and can also express a lack of evidence, i.e.\@ an ``I don't know''. We can for instance generalize the predicted output of a neural classifier using the Dirichlet distribution, allowing us to express a lack of evidence through a uniform Dirichlet.
This is different from a uniform Categorical distribution, which does not distinguish an equal probability for all classes from the lack of evidence. For the purpose of this survey, we define Evidential Deep Learning as a family of approaches in which a neural network can fall back onto a uniform prior for unknown inputs. While neural networks usually parameterize likelihood functions, approaches in this survey parameterize prior or posterior distributions instead. The advantages of this methodology are now demonstrated using the example in the following section.

\subsection{An Illustrating Example: The Iris Dataset}\label{sec:example}

\begin{figure}[t]
    \centering

    \begin{minipage}{0.15\textwidth}
        \vfill
        \begin{subfigure}[h]{\linewidth}
            \centering
            \includegraphics[width=0.9\textwidth]{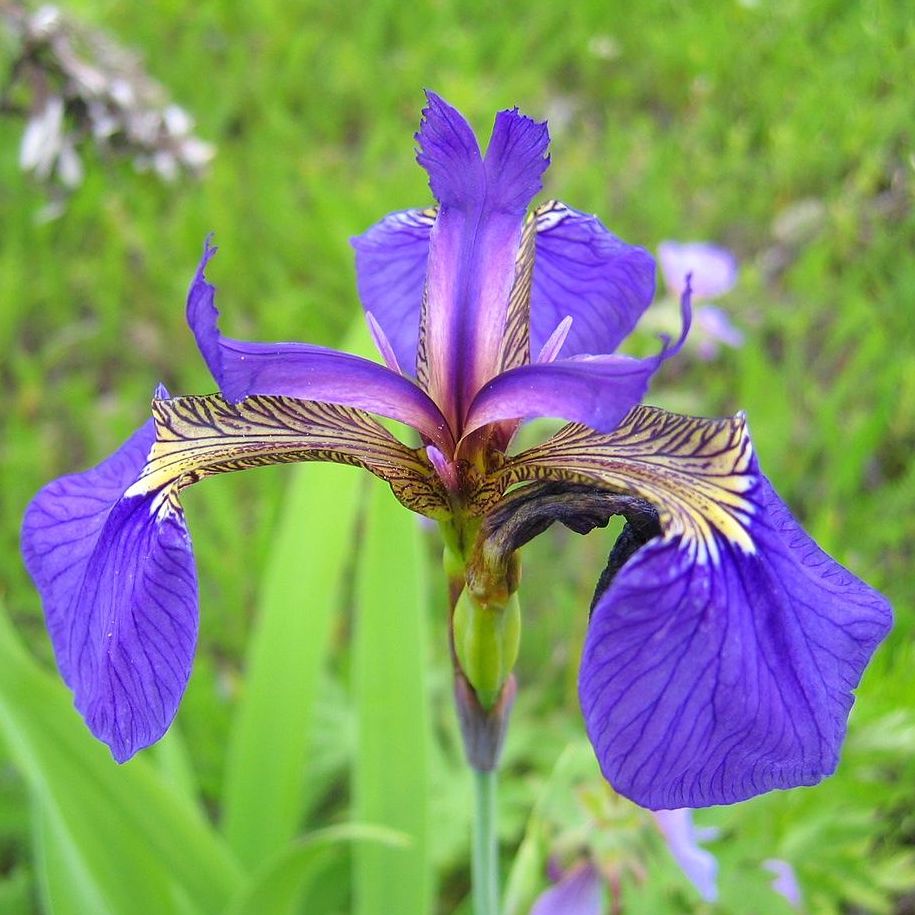}
            \caption{\emph{Iris setosa}}\label{subfig:setosa}
        \end{subfigure}
        \vfill
        \begin{subfigure}[h]{\linewidth} 
            \centering
            \includegraphics[width=0.9\textwidth]{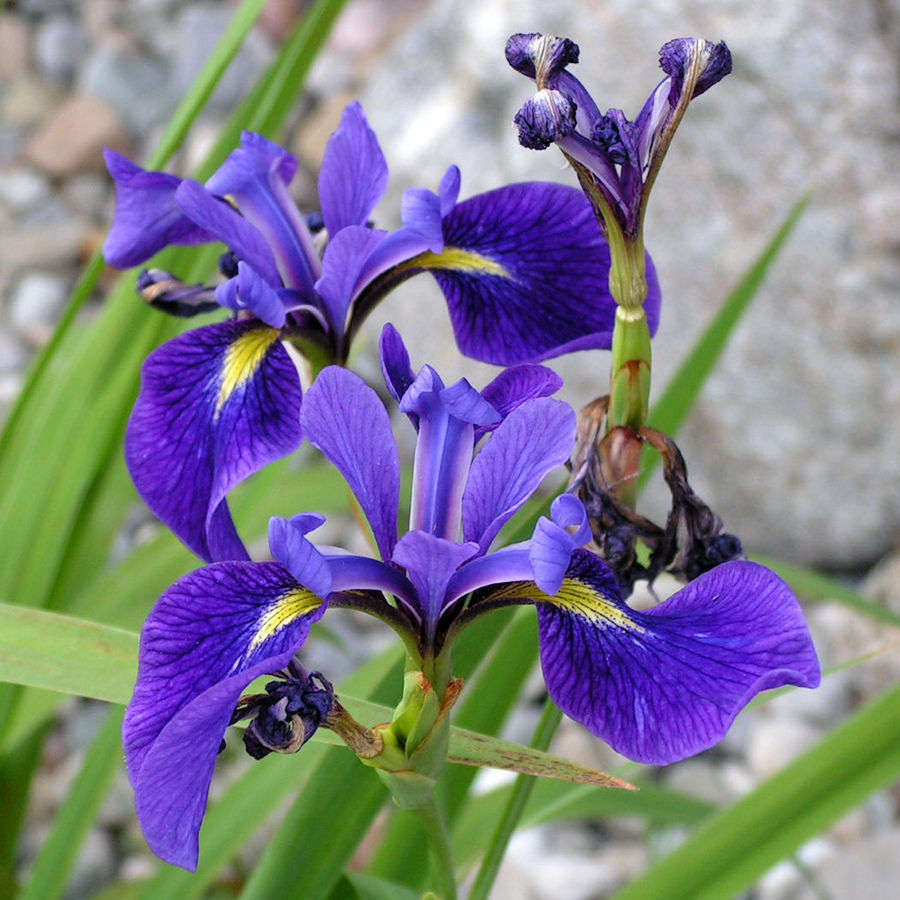}
            \caption{\emph{Iris versicolor}}\label{subfig:versicolor}
        \end{subfigure}
        \vfill
        \begin{subfigure}[h]{\linewidth} 
            \centering
            \includegraphics[width=0.9\textwidth]{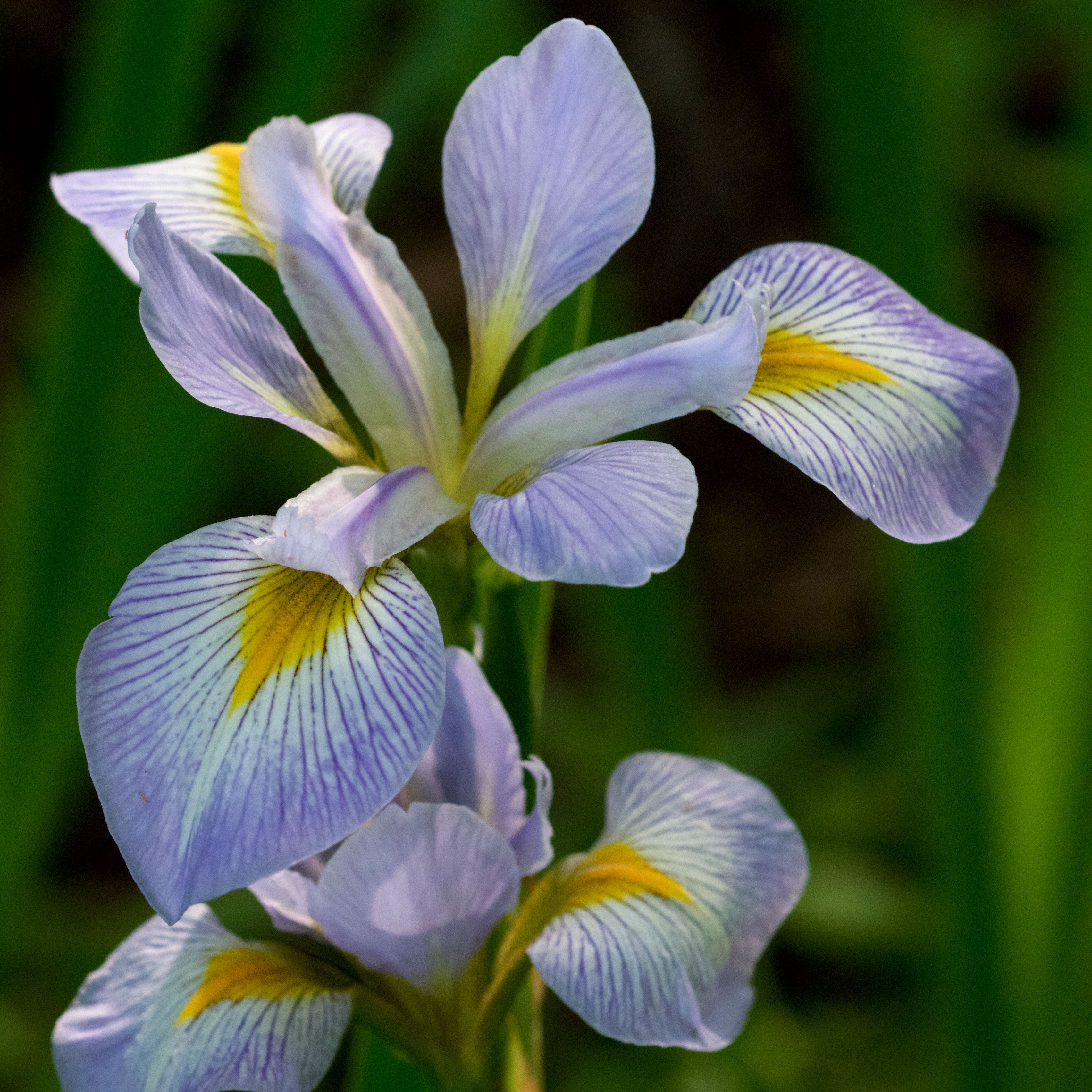}
            \caption{\emph{Iris virginica}}\label{subfig:virginica}
        \end{subfigure}
        \vfill
    \end{minipage}%
    \begin{minipage}{0.85\textwidth}
        \begin{subfigure}[t]{\linewidth}  
            \centering
            \includegraphics[width=0.975\textwidth]{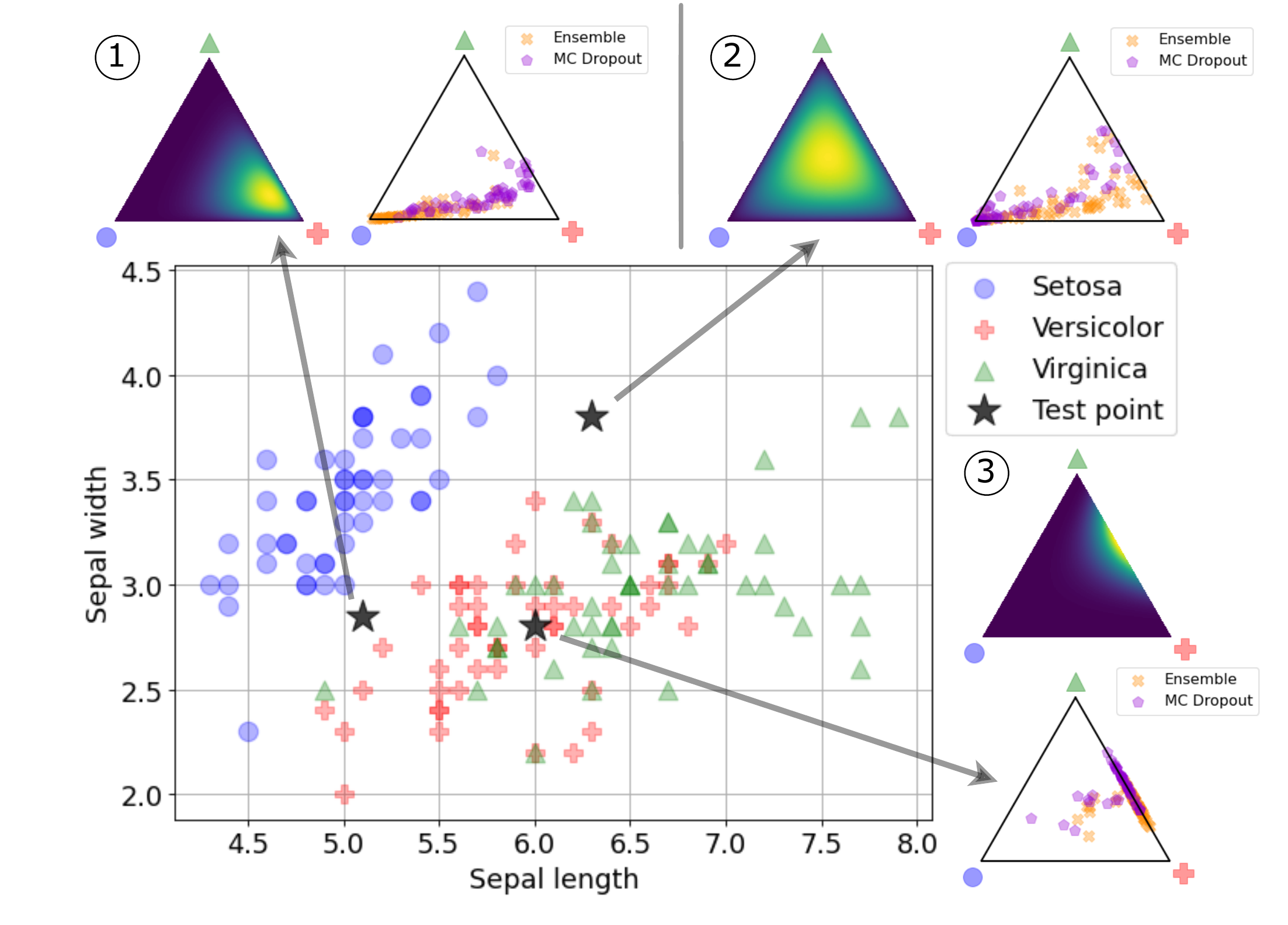}
        \end{subfigure}
    \end{minipage}

    \caption{Illustration of different approaches to uncertainty quantifying on the Iris dataset, with examples for the classes given on the left (\cref{subfig:setosa,subfig:versicolor,subfig:virginica}). On the right, the data  is plotted alongside some predictions of a prior network (lighter colors indicate higher density) and an ensemble and MC Dropout model on the probability simplex, with $50$ predictions each. Iris images were taken from \citealp{iris_setosa, iris_versicolor, iris_virginica}.}
    \label{fig:iris-example}
\end{figure}

To illustrate the advantages of EDL, we choose a classification problem based on the Iris dataset \citep{fisher1936use}. It contains measurements of three different species of iris flowers (shown in \cref{subfig:setosa,subfig:versicolor,subfig:virginica}). We use the dataset as made available through \texttt{scikit-learn} \citep{scikit-learn} and plot the relationship between the width and lengths measurements of the flowers' petals in \cref{fig:iris-example}.

We train an deep neural network ensemble \citep{lakshminarayanan2017simple} with $50$ model instances, a model with MC Dropout \citep{gal2016dropout} with $50$ predictions and a prior network \citep{sensoy2018evidential}, an example of EDL, on all available data points, and plot their predictions on three test points on the 3-probability simplex in \cref{fig:iris-example}.\footnote{For information about training and model details, see \cref{app:iris-code-details}.} On these simplices, each point signifies a Categorical distribution, with the proximity to one of the corners indicating a higher probability for the corresponding class. EDL methods for classification do not predict a single output distribution, but an entire \emph{density over output distributions}.

Test point \circled{3} lies in a region of overlap between instances of \emph{Iris versicolor} and \emph{Iris virginica}, thus inducing high aleatoric uncertainty. In this case, we can see that the prior network places all of its density on between these two classes, similar to most of the predictions of the ensemble and MC Dropout (bottom right). However, some of the latter predictions still land in the center of the simplex. The point \circled{1} is located in an area without training examples between instances of \emph{Iris versicolor} and \emph{setosa}, as well as close to a single \emph{virginica} outlier. As shown in the top left, ensemble and MC Dropout predictions agree that the point belongs to either the \emph{setosa} or \emph{versicolor} class, with a slight preference for the former. The prior network concentrates its prediction on \emph{versicolor}, but admits some uncertainty towards the two other choices. The last test point \circled{2} is placed in an area of the feature space devoid of any data, roughly equidistant from the three clusters of flowers. Similar to the previous example, the ensemble and MC dropout predictions on the top right show a preference for \emph{Iris setosa} and \emph{versicolor}, albeit with higher uncertainy. The prior network however shows an almost uniform density, admitting distributional uncertainty about this particular input.

This simple example provides some insights into the potential advantages of EDL: First of all, the prior network was able to provide reasonable uncertainty estimates in comparison with BMA methods. Secondly, the prior network is able to admit its lack of knowledge for the OOD data point by predicting an almost uniform prior, something that the other models are not able to. As laid out in \cref{sec:uncertainty-dirichlet}, EDL actually allows the user to disentangle model uncertainty due to a simple lack of data and due to the input being out-of-distribution. Lastly, training the prior network only required a single model, which is a noticeable speed-up compared to MC Dropout and especially the training of ensembles. 

\section{Evidential Deep Learning for Classification}

In order to introduce EDL methods for classification, we first give a brief introduction to the Dirichlet distribution and its role as a conjugate prior in Bayesian inference in \cref{sec:dirichlet}. We then show in \cref{sec:parameterization}
how neural networks can parameterize Dirichlet distributions, while  \cref{sec:uncertainty-dirichlet} reveals how such a parameterization can be exploited for efficient uncertainty estimation. The remaining sections enumerate different examples from the literature parameterizing either a prior (\cref{sec:prior-networks}) or posterior Dirichlet distribution (\cref{sec:posterior-networks}).

\subsection{The Dirichlet distribution}\label{sec:dirichlet}

\begin{figure}[tb]
    \centering
    \includegraphics[width=0.55\textwidth]{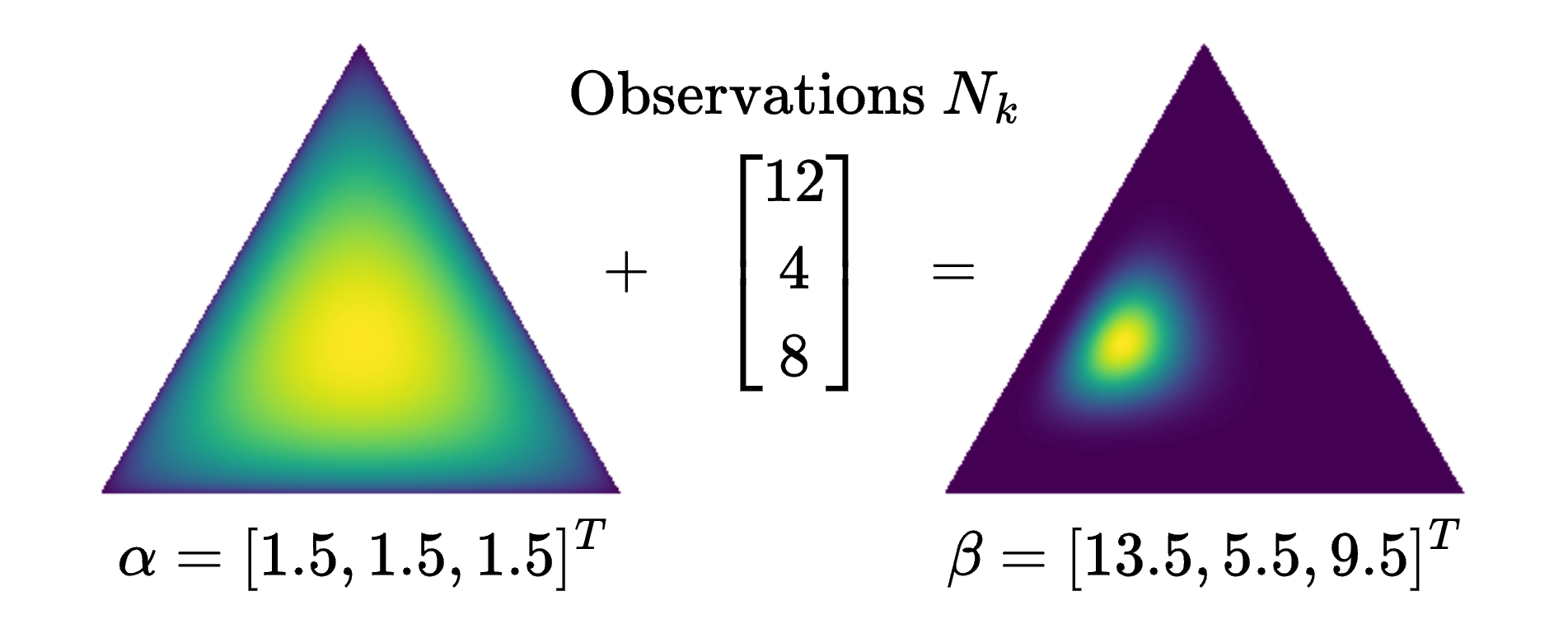}
    \caption{A prior Dirichlet distribution is updated with a vector of class observations. The posterior Dirichlet then shifts density towards the classes $k$ with more observed instances. }\label{fig:dirichlet-visualization}
\end{figure}

 Modelling for instance a binary classification problem is commonly done using the Bernoulli likelihood. The Bernoulli likelihood has a single parameter $\pi$, indicating the probability of success (or of the positive class), and is given by 
 
 \begin{equation}
     \text{Bernoulli}(y|\pi) = \pi^y(1 - \pi)^{(1 - y)}.
 \end{equation}
 
 Within Bayesian inference as introduced in \cref{sec:background}, the Beta distribution is a commonly used prior for a Bernoulli likelihood. It defines a probability distribution over the parameter $\pi$, itself possessing two shape parameters $\alpha_1$ and $\alpha_2$:

\begin{equation}\label{eq:beta}
    \text{Beta}(\pi; \alpha_1, \alpha_2) = \frac{1}{B(\alpha_1, \alpha_2)} \pi^{\alpha_1 - 1}(1 - \pi)^{\alpha_2 - 1};\quad B(\alpha_1, \alpha_2) = \frac{\Gamma(\alpha_1)\Gamma(\alpha_2)}{\Gamma(\alpha_1 + \alpha_2)} ;
\end{equation}

where $\Gamma(\cdot)$ stands for the gamma function, a generalization of the factorial to the real numbers, and $B(\cdot)$ is called the Beta function (not to be confused with the distribution). When extending the classification problem from two to an arbitrary number of classes, we use a Categorical likelihood:

\begin{equation}
    \text{Categorical}(y|\bm{\pi}) = \prod_{k=1}^K \pi_k^{\indicator{y=k}},
\end{equation}

in which $K$ denotes the number of categories or classes, and the class probabilities are expressed using a vector $\bm{\pi} \in [0, 1]^K$with $\sum_k \pi_k = 1$, and $\indicator{(\cdot)}$ is the indicator function. This distribution appears for instance in classification problems when using neural networks, since most neural networks for classification use a softmax function after their last layer to produce a Categorical distribution of classes s.t.\@ $\pi_k \equiv P(y=k|x)$. In this setting, the Dirichlet distribution arises as a suitable prior and  multivariate generalization of the Beta distribution (and is thus also called the \emph{multivariate Beta distribution}):

\begin{equation}\label{eq:conjugate-dirichlet}
    \text{Dir}(\bm{\pi}; \bm{\alpha}) = \frac{1}{B(\bm{\alpha})}\prod_{k=1}^K \pi_k^{\alpha_k-1};\quad B(\bm{\alpha}) = \frac{\prod_{k=1}^K\Gamma(\alpha_k)}{\Gamma(\alpha_0)};\quad \alpha_0 = \sum_{k=1}^K \alpha_k;\quad \alpha_k \in \mathbb{R}^+;
\end{equation}

where the Beta function $B(\cdot)$ is now defined for $K$ shape parameters compared to \cref{eq:beta}. For notational convenience, we also define $\mathbb{K} = \{1, \ldots, K\}$ as the set of all classes. The distribution is characterized by its \emph{concentration parameters} $\bm{\alpha}$, the sum of which, often denoted as $\alpha_0$, is called the \emph{precision}.\footnote{The precision is analogous to the precision of a Gaussian, where a larger $\alpha_0$ signifies a sharper distribution.} The Dirichlet is a \emph{conjugate prior} for such a Categorical likelihood, meaning that according to Bayes' rule in \cref{eq:bayes-rule}, they produce a Dirichlet posterior with parameters $\bm{\beta}$, given a data set $\mathbb{D} = \{(x_i, y_i)\}_{i=1}^N$ of $N$ observations with corresponding labels:

\begin{equation}\begin{aligned}\label{eq:posterior}
    p(\bm{\pi}|\mathbb{D}, \bm{\alpha}) & \propto p\big(\{y_i\}_{i=1}^N|\bm{\pi}, \{x_i\}_{i=1}^N\big)p(\bm{\pi}|\bm{\alpha}) = \prod_{i=1}^N\prod_{k=1}^K \pi_k^{\indicator{y_i = k}}\frac{1}{B(\bm{\alpha})}\prod_{k=1}^K \pi_k^{\alpha_k-1} \\[0.2cm] 
    & = \prod_{k=1}^K \pi_k^{\big(\sum_{i=1}^N\indicator{y_i = k}\big)}\frac{1}{B(\bm{\alpha})}\prod_{k=1}^K \pi_k^{\alpha_k-1} = \frac{1}{B(\bm{\alpha})}\prod_{k=1}^K \pi_k^{N_k + \alpha_k-1} \propto \text{Dir}(\bm{\pi}; \bm{\beta}),\\[0.2cm] 
\end{aligned}\end{equation}

where $\bm{\beta}$ is a vector with $\beta_k = \alpha_k + N_k$, with $N_k$ denoting the number of observations for class $k$. Intuitively, this implies that the prior belief encoded by the initial Dirichlet is updated using the actual data, sharpening the distribution for classes for which many instances have been observed. Similar to the Beta distribution in \cref{eq:beta}, the Dirichlet is a \emph{distribution over Categorical distributions} on the $K-1$ probability simplex; we show an example with its concentration parameters and the Bayesian update in \cref{fig:dirichlet-visualization}. 

\subsection{Parameterization}\label{sec:parameterization}

For a classification problem with $K$ classes, a neural classifier is usually realized as a function $f_{\bm{\theta}}: \mathbb{R}^D \rightarrow \mathbb{R}^K$, mapping an input $\bx \in \mathbb{R}^D$ to \emph{logits} for each class. Followed by a softmax function, this then defines a Categorical distribution over classes with a vector $\bm{\pi}$ with $\pi_k \equiv p(y=k|\bx, \bm{\theta})$. The same underlying architecture can be used without any major modification to instead parameterize a \emph{Dirichlet} distribution, predicting a distribution \emph{over Categorical distributions} $p(\bm{\pi}|\bx, \hat{\bm{\theta}})$ as in \cref{eq:conjugate-dirichlet}.\footnote{The only thing to note here is that the every $\alpha_k$ has to be strictly positive, which can for instance be enforced by using an additional softplus, exponential or ReLU function \citep{sensoy2018evidential, malinin2018predictive, sensoy2020uncertainty}.} In order to classify a data point $\bx$, a Categorical distribution is created from the predicted concentration parameters of the Dirichlet as follows (this corresponds to the mean of the Dirichlet, see  \cref{app:expectation-dirichlet}):

\begin{equation}
    \bm{\alpha} = \exp\big(f_{\bm{\theta}}(\bx)\big);\quad \pi_k = \frac{\alpha_k}{\alpha_0};\quad \hat{y} = \argmax_{k \in \mathbb{K}}\ \pi_1, \ldots, \pi_K .
\end{equation}

Parameterizing a Dirichlet posterior distribution follows a similar logic, as we will discuss in \cref{sec:posterior-networks}.

\subsection{Uncertainty Estimation with Dirichlet Networks}\label{sec:uncertainty-dirichlet}

\begin{figure}[tb]
    \centering
    \begin{subfigure}[t]{0.3\linewidth}
        \centering
        \includegraphics[width=0.95\linewidth]{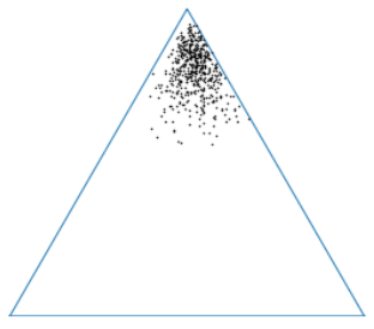}
        \caption{Categorical distributions predicted by a neural ensemble on the probability simplex.}
        \label{subfig:simplex-ensemble}
    \end{subfigure}
    \hfill
    \begin{subfigure}[t]{0.3\linewidth}
        \centering
        \includegraphics[width=0.95\linewidth]{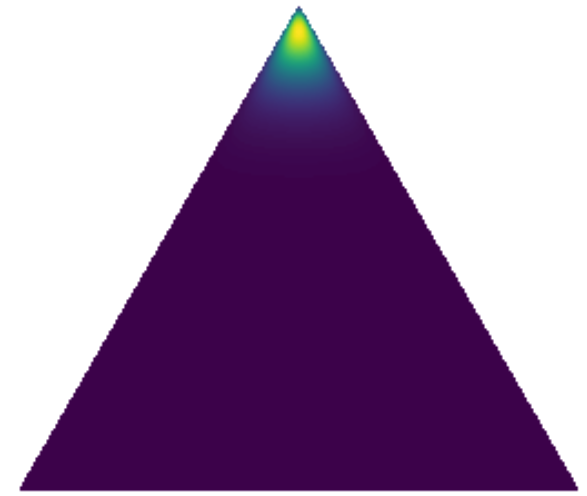}
        \caption{Probability simplex for a confident prediction, for with the density concentrated in a single corner.}
        \label{subfig:simplex-confident}
    \end{subfigure}
    \hfill
    \begin{subfigure}[t]{0.3\linewidth}
        \centering
        \includegraphics[width=0.95\linewidth]{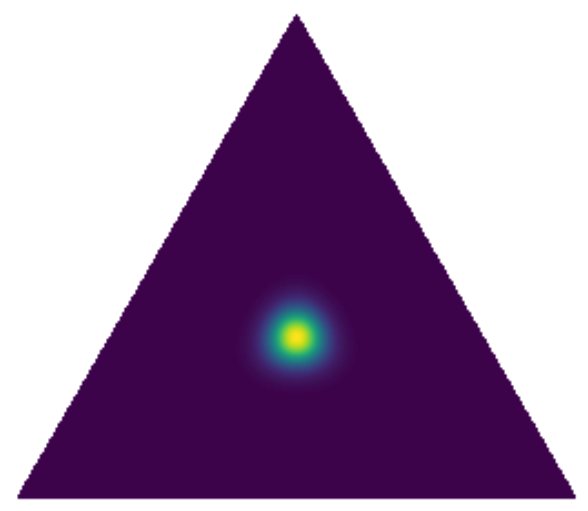}
        \caption{Dirichlet distribution for a case of data uncertainty, with the density concentrated in the center.}
        \label{subfig:simplex-aleatoric}
    \end{subfigure}
    
    \bigskip
    \begin{subfigure}[t]{0.3\linewidth}
        \centering
        \includegraphics[width=0.95\linewidth]{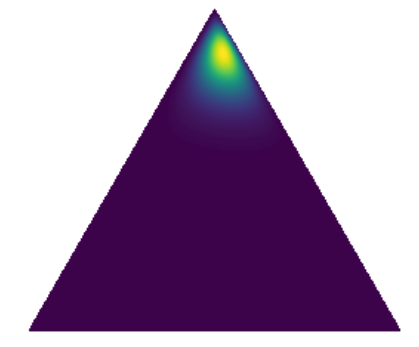}
        \caption{Dirichlet distribution for a case of model uncertainty, with the density spread out more.}
        \label{subfig:simplex-epistemic}
    \end{subfigure}
    \hfill
    \begin{subfigure}[t]{0.3\linewidth}
        \centering
        \includegraphics[width=0.95\linewidth]{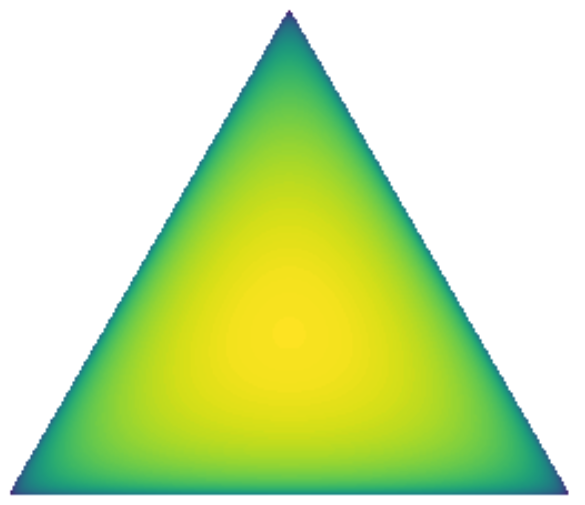}
        \caption{Dirichlet for a case of distributional uncertainty, with the density spread across the whole simplex.}
        \label{subfig:simplex-distributional}
    \end{subfigure}
    \hfill
    \begin{subfigure}[t]{0.3\linewidth}
        \centering
        \includegraphics[width=0.95\linewidth]{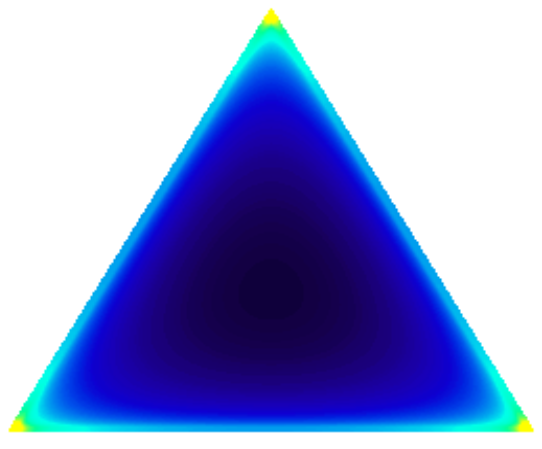}
        \caption{Alternative approach to distributional uncertainty called representation gap, with density concentrated along the edges.}
        \label{subfig:simplex-representation-gap}
    \end{subfigure}
    \caption{Examples of the probability simplex for a $K=3$ classification problem, where every corner corresponds to a class and every point to a Categorical distribution. Brighter colors correspond to higher density. (a) Predicted Categorical distributions by an ensemble of discriminators. (b) -- (e) (Desired) Behavior of Dirichlet in different scenarios by \citet{malinin2018predictive}: (b) For a confident prediction, the density is concentrated in the corner of the simplex corresponding to the assumed class. (c) In the case of aleatoric uncertainty, the density is concentrated in the center, and thus uniform Categorical distributions are most likely. (d) In the case of model uncertainty, the density may still be concentrated in a corner, but more spread out, expressing the uncertainty about the right prediction. (e) In the case of an OOD input, a uniform Dirichlet expresses that any Categorical distribution is equally likely, since there is no evidence for any known class.  (f) Representation gap by \citet{nandy2020towards}, proposed as an alternative behavior for OOD data. Here, the density is instead concentrated solely on the edges of the simplex.}
    \label{fig:simplex}
\end{figure}

Let us now turn our attention to how to estimate the aleatoric, epistemic and distributional uncertainty as laid out in \cref{sec:uncertainty-quantification} within the Dirichlet framework. In \cref{fig:simplex}, we show different shapes of a Dirichlet distribution parameterized by a neural network, corresponding to different cases of uncertainty, where each point on the simplex represents a Categorical distribution, with proximity to a corner indicating a high probability for the corresponding class. \cref{subfig:simplex-ensemble} displays the predictions of an ensemble of classifiers as a point cloud on the simplex. Using a Dirichlet, this finite set of distributions can be extended to a continuous density over the whole simplex. As we will see in the following sections, parameterizing a Dirichlet distribution with a neural network enables us to distinguish different scenarios using the shape of its density, as shown in \cref{subfig:simplex-confident,subfig:simplex-aleatoric,subfig:simplex-epistemic,subfig:simplex-distributional,subfig:simplex-representation-gap}, which we will discuss in more detail along the way.

However, since we do not want to inspect Dirichlets visually, we instead use closed form expression to quantify uncertainty, which we will discuss now. Although stated for the prior parameters $\bm{\alpha}$, the following methods can also be applied to the posterior parameters $\bm{\beta}$ without loss of generality.

\paragraph{Data (aleatoric) uncertainty} To obtain a measure of data uncertainty, we can evaluate the expected entropy of the data distribution $p(y|\bm{\pi})$ (similar to previous works like e.g.\@ \citeauthor{gal2016dropout}, \citeyear{gal2016dropout}). As the entropy captures the ``peakiness'' of the output distribution, a lower entropy indicates that the model is concentrating most probability mass on a single class, while high entropy characterizes a more uniform distribution---the model is undecided about the right prediction. For Dirichlet networks, this quantity has a closed-form solution (for the full derivation, refer to \cref{app:expected-entropy}):

\begin{equation}\label{eq:expected-entropy}
    \Expect[\bigg]{p(\bm{\pi}|\bx, \hat{\bm{\theta}})}{H\Big[P(y|\bm{\pi})\Big]} = - \sum_{k=1}^K\frac{\alpha_k}{\alpha_0}\bigg(\psi(\alpha_k+1) -  \psi(\alpha_0+1)\bigg) 
\end{equation}

where $\psi$ denotes the digamma function, defined as $\psi(x) = \frac{d}{d x} \log \Gamma(x)$, and $H$ the Shannon entropy.

\paragraph{Model (epistemic) uncertainty} 
As we saw in \cref{sec:uncertainty-quantification}, most approaches in the Dirichlet framework avoid the intractable integral over network parameters $\btheta$ by using a point estimate $\hat{\btheta}$.\footnote{With exceptions such as \citet{haussmann2019bayesian,zhao2020uncertainty}. When the distribution over parameters in  \cref{eq:prior-networks-fac} is retained, alternate expressions of the aleatoric and epistemic uncertainty are derived by \citet{woo2022analytic}.} This means that computing the model uncertainty via the weight posterior $p(\bm{\theta}|\mathbb{D})$ like in \citet{blundell2015weight, gal2016dropout, gal2018understanding} is not possible.  Nevertheless, a key property of Dirichlet networks is that epistemic uncertainty is expressed in the spread of the Dirichlet distribution (for instance in \cref{fig:simplex} (d) and (e)). Therefore, the epistemic uncertainty can be quantified considering the concentration parameters $\bm{\alpha}$ that shape this distribution: \citet{charpentier2020posterior} simply consider the maximum $\alpha_k$ as a score akin to the maximum probability score by \citet{hendrycks2017baseline}, while \citet{sensoy2018evidential} compute it by $K / \sum_{k=1}^K (\alpha_k + 1)$ or simply $\alpha_0$ \citep{charpentier2020posterior}. In both cases, the underlying intuition is that larger $\alpha_k$ produce a sharper density, and thus indicate increased confidence in a prediction.
 
\paragraph{Distributional uncertainty} Another appealing property of this model family is being able to distinguish uncertainty due to model underspecification (\cref{subfig:simplex-epistemic}) from uncertainty due to unknown inputs (\cref{subfig:simplex-distributional}). In the Dirichlet framework, the distributional uncertainty can be quantified by computing the difference between the total amount of uncertainty and the data uncertainty, which can be expressed through the mutual information between the label $y$ and its Categorical distribution $\bm{\pi}$:

\begin{equation}\label{eq:dirichlet-mi}
    I\Big[y, \bm{\pi}\Big| \bx, \mathbb{D}\Big] = \underbrace{H\bigg[\Expect[\Big]{p(\bm{\pi}|\bx, \mathbb{D})}{P(y|\bm{\pi})}\bigg]}_{\text{Total Uncertainty}} - \underbrace{\Expect[\bigg]{p(\bm{\pi}|\bx, \mathbb{D})}{H\Big[P(y|\bm{\pi})\Big]}}_{\text{Data Uncertainty}}
\end{equation}

This quantity expresses how much information we would receive about $\bpi$ if we were given the label $y$, conditioned on the new input $\bx$ and the training data $\mathbb{D}$. In regions in which the model is well-defined, receiving $y$ should not provide much new information about $\bpi$---and thus the mutual information would be low. Yet, such knowledge should be very informative in regions in which few data have been observed, and there this mutual information would indicate higher distributional uncertainty.
Given that $\Expect[]{}{\pi_k} = \frac{\alpha_k}{\alpha_0}$ (\cref{app:expectation-dirichlet}) and assuming the point estimate $p(\bm{\pi}|\bx, \mathbb{D}) \approx p(\bm{\pi}|\bx, \hat{\bm{\theta}})$ to be sufficient \citep{malinin2018predictive}, we obtain an expression very similar to \cref{eq:expected-entropy}:

\begin{equation}
    I\Big[y, \bm{\pi}\Big| \bx, \mathbb{D}\Big] = - \sum_{k=1}^K \frac{\alpha_k}{\alpha_0}\bigg(\log \frac{\alpha_k}{\alpha_0} -\psi(\alpha_k+1) + \psi(\alpha_0+1)\bigg) 
\end{equation}

\paragraph{Note on epistemic uncertainty estimation} The introduction of distributional uncertainty, a notion that is non-existent in the Bayesian Model Averaging framework, warrants a note on the estimation of epistemic uncertainty in general. Firstly, since we often use the point estimate $ p(\btheta|\mathbb{D}) \approx \delta(\btheta - \hat{\btheta})$ from \cref{eq:prior-networks-fac} in Evidential Deep Learning, model uncertainty usually is no longer estimated via the uncertainty in the weight posterior, but instead through the parameters of the prior or posterior distribution. Furthermore, even though they appear similar, distributional uncertainty is different from epistemic uncertainty, since it is the uncertainty in the distribution $p(\bpi|\bx, \btheta)$. Distinguishing epistemic from distributional uncertainty also allows us to differentiate uncertainty due to underspecification from uncertainty due to a lack of evidence. In BMA, these notions are indistinguishable: In theory, model uncertainty on OOD data should be high since the model is underspecified on them, however theoretical and empirical work has shown this is not always the case \citep{ulmer2020trust, ulmer2021know, van2022benchmarking}. Even then, the additive decomposition of the mutual information has been critized since the model will also have a great deal of \emph{uncertainty about its aleatoric uncertainty} in the beginning of the training process \citep{hullermeier2022quantifying}, and thus this decomposition might not be accurate. Furthermore, even when we obtain the best possible model within its hypothesis class, using the discussed methods it is impossible to estimate uncertainty induced by a misspecified hypothesis class. This can motivate approaches in which a second, auxiliary model directly predicts model uncertainty of a target model  \citep{jain2021deup, zerva2022better}. 

\subsection{Existing Approaches for Dirichlet Networks}\label{sec:approaches}

Being able to quantify aleatoric, epistemic and distributional uncertainty in a single forward pass and in closed form are desirable traits, as they simplify the process of obtaining different uncertainty scores. However, it is important to note that the behavior of the Dirichlet distributions in \cref{fig:simplex} is idealized. In the usual way of training neural networks through empirical risk minimization, Dirichlet networks are not incentivized to behave in the depicted way. Thus, when comparing existing approaches for parameterizing Dirichlet priors in \cref{sec:prior-networks} and posteriors in \cref{sec:posterior-networks},\footnote{Even though the term \emph{prior} and \emph{posterior network} were coined by \citet{malinin2018predictive} and \citet{charpentier2020posterior} for their respective approaches, we use them in the following as an umbrella term for all methods targeting a prior or posterior distribution.} we mainly focus on the different ways in which authors try to tackle this problem by means of loss functions and training procedures. We give an overview over the discussed works in \cref{tab:overview-prior,tab:overview-posterior} in these respective sections. For additional details, we refer the reader to \cref{app:fundamental-derivations} for general derivations concerning the Dirichlet distribution. We dedicate \cref{app:additional-derivations} to derivations of the different loss functions and regularizers and give a detailed overview over their mathematical forms in \cref{app:overview-loss-functions}. Available code repositories for all works surveyed are listed in \cref{app:code-availability}.

\subsubsection{Prior Networks}\label{sec:prior-networks}

\begin{table}[tb]
    \centering
    \caption{Overview over prior networks for classification. $(*)$ OOD samples were created inspired by the approach of \citet{liang2018enhancing}. ID: Using in-distribution data samples.}
     \resizebox{0.95\textwidth}{!}{
        \renewcommand{\arraystretch}{1.6}
        \begin{tabular}{@{}llll@{}}
            \toprule
            Method & Loss function & Architecture & \thead[tl]{OOD-free\\training?} \\
            \midrule
            \makecell[tl]{Prior network\\ \citep{malinin2018predictive}}       &  \makecell[tl]{ID KL w.r.t smoothed label \&\\OOD KL w.r.t. uniform prior} & MLP / CNN & \xmark \\
            \makecell[tl]{Prior networks\\ \citep{malinin2019reverse}}         &  \makecell[tl]{Reverse KL of  \citet{malinin2018predictive}} & CNN & \xmark \\
            \makecell[tl]{Information Robust Dirichlet Networks\\ \citep{tsiligkaridis2019information}} & \makecell[tl]{$l_p$ norm w.r.t one-hot label \& \\Approx. R\'enyi divergence\\ w.r.t. uniform prior} & CNN & \cmark \\
            \makecell[tl]{Dirichlet via Function Decomposition\\ \citep{bilovs2019uncertainty}} &  \makecell[tl]{Uncertainty Cross-entropy \&\\mean \& variance regularizer} & RNN & \cmark \\ 
            \makecell[tl]{Prior network with PAC Regularization \\ \citep{haussmann2019bayesian}} &  \makecell[tl]{Negative log-likelihood loss +\\ PAC regularizer} & BNN & \cmark \\
            \makecell[tl]{Ensemble Distribution  Distillation\\ \citep{malinin2020ensemble}} & Knowledge distillation objective & MLP / CNN & \cmark \\
            \makecell[tl]{Self-Distribution Distillation\\ \citep{fathullah2022self}} & Knowledge distillation objective & CNN & \cmark \\
            \makecell[tl]{Prior networks with representation gap\\ \citep{nandy2020towards}} &  \makecell[tl]{ID \& OOD Cross-entropy + \\ precision regularizer} & MLP / CNN & \xmark \\
            \makecell[tl]{Prior RNN  \citep{shen2020modeling}} & Cross-entropy + entropy regularizer & RNN & (\xmark)$^*$  \\
            \makecell[tl]{Graph-based Kernel Dirichlet distribution\\estimation (GKDE)
            \citep{zhao2020uncertainty}} &  \makecell[tl]{$l_2$ norm w.r.t. one-hot label \& \\ KL reg. with node-level distance prior \&\\ Knowledge distillation objective} & GNN & \cmark \\
            \bottomrule
        \end{tabular}%
    }
    \label{tab:overview-prior}
\end{table}

The key challenge in training Dirichlet networks is to ensure both high classification performance and the intended behavior under OOD inputs. For this reason, most discussed works follow a loss function design using two parts: One optimizing for task accuracy to achieve the former goal, the other optimizing for a flat Dirichlet distribution, as flatness suggests a lack of evidence. To enforce flatness, the predicted Dirichlet is compared to a uniform distribution using some probabilistic divergence measure. We divide prior networks into two groups: Approaches using additional OOD data for this purpose (\emph{OOD-dependent approaches}), and those which do not required OOD data (\emph{OOD-free approaches}), as listed in \cref{tab:overview-prior}. 


\paragraph{OOD-free approaches} Apart from a standard negative log-likelihood loss (NLL) as used by \citet{haussmann2019bayesian}, one simple approach to optimizing the model is to impose a $l_p$-loss between the one-hot encoding $\by$ of the original label $y$ and the Categorical distribution $\bpi$. \citet{tsiligkaridis2019information} show that since the values of $\bpi$ depend directly on the predicted concentration parameters $\balpha$, a generalized loss can be derived to be upper-bounded by the following expression (see the full derivation given in \cref{app:infinity-norm-loss}):

\begin{equation}
    \Expect[\big]{p(\bm{\pi}|\bx, \bm{\theta})}{||\by - \bm{\pi}||_p} \le  \bigg(\frac{\Gamma(\alpha_0)}{\Gamma(\alpha_0 + p)}\bigg)^\frac{1}{p}\vast(\frac{\Gamma\Big(\sum_{k \neq y}\alpha_k + p\Big)}{\Gamma\Big(\sum_{k\neq y} \alpha_k\Big)} + \sum_{k \neq y}\frac{\Gamma(\alpha_k + p)}{\Gamma(\alpha_k)} \vast)^\frac{1}{p} 
\end{equation}

Since the sum over concentration parameters excludes the one corresponding to the true label, this loss can be seen as reducing the density on the areas of the probability simplex that do not correspond to the target class. \citet{sensoy2018evidential} specifically utilize the $l_2$ loss, which has the following form (see \cref{app:il2-norm-loss}):

\begin{equation}
     \Expect[\Big]{p(\bm{\pi}|\bx, \bm{\theta})}{||\by - \bm{\pi}||_2^2} =  \sum_{k=1}^K \Big(\indicator{y = k} -\frac{\alpha_k}{\alpha_0}\Big)^2 + \frac{\alpha_k(\alpha_0 - \alpha_k)}{\alpha_0^2(\alpha_0 + 1)}
\end{equation}

where $\indicator{(\cdot)}$ denotes the indicator function. Since $\alpha_k / \alpha_0 \le 1$, we can see that the term with the indicator functions penalizes the network when the concentration parameter $\alpha_k$ corresponding to the correct label does not exceed the others. The remaining aspect lies in the regularization: To achieve reliable predictive uncertainty, the density associated with incorrect classes should be reduced. One such option is to decrease the Kullback-Leibler divergence from a uniform Dirichlet (see \cref{app:kl-dirichlets}):

\begin{equation}
    \text{KL}\Big[p(\bm{\pi}|\bm{\alpha})\Big|\Big| p(\bm{\pi}|\bm{1})\Big] = \log \frac{\Gamma(K)}{B(\bm{\alpha})} + \sum_{k=1}^K (\alpha_k - 1)\big(\psi(\alpha_k) - \psi(\alpha_0)\big)
\end{equation}

In the case of \citet{zhao2020uncertainty}, who apply their model to graph structures, they do not decrease the divergence from a uniform Dirichlet, but incorporate information about the local graph neighborhood into the reference distribution by considering the distance from  and label of close nodes.\footnote{They also add another knowledge distillation term \citep{hinton2015distilling} to their loss, for which the model tries to imitate the predictions of a vanilla Graph Neural Network that functions as the teacher network.} Nevertheless, the KL-divergence w.r.t. a uniform Dirichlet is used by many of the following works. Other divergence measures are also possible: \citet{tsiligkaridis2019information} instead use a local approximation of the R\'enyi divergence.\footnote{The Kullback-Leibler divergence can be seen as a special case of the R\'enyi divergence \citep{erven2014renyi}, where the latter has a stronger information-theoretic underpinning.} First, the concentration parameter for the correct class $\alpha_y$ is removed from the Dirichlet by creating $\tilde{\balpha} = (1 - \by) \cdot \balpha + \by$. Then, the remaining concentration parameters are pushed towards uniformity by the divergence measure, which can be derived to be

\begin{equation}
    \text{R\'enyi}\Big[p(\bm{\pi}|\tilde{\bm{\alpha}})\Big|\Big| p(\bm{\pi}|\bm{1})\Big] \approx \frac{1}{2}\Big[\sum_{k\neq y} \big(\alpha_k -1\big)^2\big(\psi^{(1)}(\alpha_j) - \psi^{(1)}(\tilde{\alpha}_0)\big) - \psi^{(1)}(\tilde{\alpha}_0)\sum_{\substack{k\neq k^\prime\\k\neq y,\ k^\prime\neq y}}\big(\alpha_k - 1\big)\big(\alpha_{k^\prime} - 1\big) \Big]
\end{equation}

where $\psi^{(1)}$ denotes the first-order polygamma function, defined as $\psi^{(1)}(x) = \frac{d}{d x} \psi(x)$. Since the sums ignore the concentration parameter of the correct class, only the ones of the incorrect classes are penalized.
 \citet{haussmann2019bayesian} derive an entirely different regularizer using Probably Approximately Correct (PAC) bounds from learning theory, that together with the negative log-likelihood gives a proven bound to the expected true risk of the classifier. Setting a scalar $\delta$ allows one to set the desired risk, i.e.\@ the model's expected risk is guaranteed to be the same or less than the derived PAC bound with a probability of $1 - \delta$. For a problem with $N$ available training data points, the following \emph{upper bound} is presented:
 
 \begin{equation}
    \sqrt{\frac{ \text{KL}\big[p(\bm{\pi}|\bm{\alpha})\big|\big|p(\bm{\pi}|\mathbf{1})\big] - \log \delta}{N} - 1}.
 \end{equation}

 This upper bound is then used as the actual regularizer term in practice.
  We see that even from the learning-theoretic perspective, this method follows the intuition of the original KL regularizer in a shifted and scaled form. \citet{haussmann2019bayesian} also admit that in this form, the regularizer does not allow for a direct PAC interpretation anymore, since its approximates only admits a loose bound on the risk. Yet, they demonstrate its usefulness in their experiments. Summarizing all of the presented approaches thus far, we can see that they try to force the model to concentrate the Dirichlet's density solely on the parameter corresponding to the right label---expecting a more flat density for difficult or unknown inputs. 

\paragraph{Knowledge distillation} A way to avoid the use of OOD examples while still using external information for regularization is to use \emph{knowledge distillation} \citep{hinton2015distilling}. Here, the core idea lies in a student model learning to imitate the predictions of a more complex teacher model. \citet{malinin2020ensemble} exploit this idea and show that prior networks can also be distilled using an ensemble of classifiers and their predicted Categorical distributions (akin to learning \cref{subfig:simplex-distributional} from \cref{subfig:simplex-ensemble}), which does not require regularization at all, but comes at the cost of having to train an entire ensemble a priori. Trying to solve this shortcoming, \citet{fathullah2022self} propose to use a shared feature extractor between the student and the teacher network. Instead of training an ensemble, diverse predictions are obtained from the teacher network through the use of Gaussian dropout, which are distilled into a Dirichlet distribution as in \citet{malinin2020ensemble}.  

\paragraph{OOD-dependent approaches} A uniform Dirichlet in the face of unknown inputs can also be achieved explicitly by training with OOD inputs and learning to be uncertain on them. We discuss a series of works utilizing this direction next. \citet{malinin2018predictive} simply minimize the KL divergence to a uniform Dirichlet on OOD data points. This way, the model is encouraged to be agnostic about its prediction in the face of unknown inputs. Further, instead of an $l_p$ norm, they utilize another KL term to train the model on predicting the correct label, minimizing the distance between the predicted concentration parameters and the true label. However, since only a gold \emph{label} and not a gold \emph{distribution} is available, they create one by re-distributing some of the density from the correct class onto the rest of the simplex (see \cref{app:overview-loss-functions} for full form). In their follow-up work, \citet{malinin2019reverse} argue that the asymmetry of the KL divergence as the main objective creates undesirable properties in producing the correct behavior of the predicted Dirichlet, since it creates a multi- instead of unimodal target distribution. They therefore propose to use the reverse KL instead (see \cref{app:reverse-kl-loss} for the derivation), which enforces the desired unimodal target. \citet{nandy2020towards} refine this idea further, stating that even with reverse KL training high epistemic and high distributional uncertainty (\cref{subfig:simplex-epistemic,subfig:simplex-distributional}) might be confused, and instead propose novel loss functions producing a \emph{representation gap} (\cref{subfig:simplex-representation-gap}), which aims to be more easily distinguishable. In this case, spread out densities signify epistemic uncertainty, whereas densities concentrated entirely on the edges of the simplex indicate distributional uncertainty. The way they achieve this goal is two-fold: In addition to minimizing the negative log-likelihood loss on in-domain and maximizing the entropy on OOD examples, they also penalize the precision of the Dirichlet (see \cref{app:overview-loss-functions} for full form). Maximizing the entropy on OOD examples hereby serves the same function as minimizing the KL w.r.t. to a uniform distribution, and can be implemented using the closed-form solution in \cref{app:entropy-dirichlet}:

\begin{equation}
    H\big[p(\bm{\pi}|\balpha)\big] = \log B(\bm{\alpha}) + (\alpha_0 - K)\psi(\alpha_0) - \sum_{k=1}^K (\alpha_k - 1)\psi(\alpha_k) 
\end{equation}

\paragraph{Sequential models} We also have identified two sequential applications of prior networks in the literature: For Natural Language Processing, \citet{shen2020modeling} train a recurrent neural network for spoken language understanding using a simple cross-entropy loss. Instead of using OOD examples for training, they  maximize the entropy of the model on data inputs given a learned, noisy version of the predicted concentration parameters. In comparison, \citet{bilovs2019uncertainty} apply their model to asynchronous event classification and note that the standard cross-entropy loss only involves a point estimate of a Categorical distribution, discarding all the information contained in the predicted Dirichlet. For this reason, they propose an \emph{uncertainty-aware} cross-entropy (UCE) loss instead, which has a closed-form solution in the Dirichlet case (see \cref{app:uce-loss})

\begin{equation}
    \mathcal{L}_\text{UCE} =  \psi(\alpha_y) - \psi(\alpha_0),
\end{equation}

with $\psi$ referring to the digamma function. By mimizing the difference between the digamma values of $\alpha_y$ and $\alpha_0$, the model learns to concentrate density on the correct class. Since their final concentration parameters are created using additional information from a class-specific Gaussian process, they further regularize the mean and variance for OOD data points using an extra loss term, incentivizing a loss mean and a variance corresponding to a pre-defined hyperparameter.

\subsubsection{Posterior Networks}\label{sec:posterior-networks}

\begin{table}[tb]
\caption{Overview over posterior networks for classification. OOD data is created using $(\dagger)$ the fast-sign gradient method \citep{kurakin2017adversarial}, a $(\ddagger)$ Variational Auto-Encoder (VAE; \citeauthor{kingma2014autoencoding}, \citeyear{kingma2014autoencoding}) or $(\mathsection)$ a Wasserstein GAN (WGAN; \citeauthor{arjovsky2017wasserstein}, \citeyear{arjovsky2017wasserstein}). NLL: Negative log-likelihood. CE: Cross-entropy.}
\centering
\resizebox{0.95\textwidth}{!}{
        \renewcommand{\arraystretch}{1.6}
        \begin{tabular}{@{}llll@{}}
            \toprule
            Method & Loss function & Architecture & \thead[tl]{OOD-free\\training?} \\
            \midrule
            \makecell[tl]{Evidential Deep Learning\\  \citep{sensoy2018evidential}} &  \makecell[tl]{$l_2$ norm w.r.t.\@ one-hot label + \\ KL w.r.t. uniform prior} & CNN & \cmark \\
            \makecell[tl]{Regularized ENN\\ \citep{zhao2019quantifying}} & \makecell[tl]{$l_2$ norm w.r.t.\@ one-hot label + \\ Uncertainty regularizer on OOD/ difficult samples} & MLP / CNN & \xmark \\
            \makecell[tl]{WGAN--ENN\\ \citep{hu2021multidimensional}} & \makecell[tl]{$l_2$ norm w.r.t.\@ one-hot label + \\ Uncertainty regularizer on synth. OOD} & \makecell[tl]{MLP / CNN + \\ WGAN} &  (\xmark)$^\mathsection$ \\
            \makecell[tl]{Variational Dirichlet\\ \citep{chen2018variational}} &  \makecell[tl]{ELBO +\\ Contrastive Adversarial Loss} & CNN & (\xmark)$^\dagger$ \\
            \makecell[tl]{Dirichlet Meta-Model\\ \citep{shen2022post}} & \makecell[tl]{ELBO + \\ KL w.r.t. uniform prior} & CNN & \cmark \\
            \makecell[tl]{Belief Matching  \citep{joo2020being}}           &  \makecell[tl]{ELBO} & CNN & \cmark \\ 
            \makecell[tl]{Posterior Networks\\ \citep{charpentier2020posterior}} &  \makecell[tl]{Uncertainty CE \citep{bilovs2019uncertainty} \\ + Entropy regularizer} & \makecell[tl]{MLP / CNN + \\ Norm. Flow} & \cmark \\
            \makecell[tl]{Graph Posterior Networks\\ \citep{stadler2021graph}} & Same as \citet{charpentier2020posterior} & GNN & \cmark \\
            \makecell[tl]{Generative Evidential Neural Networks\\ \citep{sensoy2020uncertainty}} & \makecell[tl]{Contrastive NLL + KL between\\ uniform \& Dirichlet of wrong classes} & CNN & (\xmark)$^\ddagger$ \\
            \bottomrule
        \end{tabular}%
    }
    \label{tab:overview-posterior}
\end{table}

As elaborated on in \cref{sec:dirichlet}, choosing a Dirichlet prior, due to its conjugacy to the Categorical distribution, induces a Dirichlet posterior distribution. Like the prior before, surveyed works listed in \cref{tab:overview-posterior}  parameterize the posterior with a neural network.  The challenges hereby are two-fold: Accounting for the number of class observations $N_k$ that make up part of the posterior density parameters $\bm{\beta}$ (\cref{eq:posterior}), and, similarly to prior networks, ensuring the wanted behavior on the probability simplex for in- and out-of-distribution inputs. 
\citet{sensoy2018evidential} base their approach on the Dempster-Shafer theory of evidence (\citealp{yager2008classic}; lending its name to the term ``Evidential Deep Learning'')  and its formalization via subjective logic \citep{audun2018subjective}, where subjective beliefs about probabilities are expressed through Dirichlet distributions. In doing so, an agnostic belief in form of a uniform Dirichlet prior $\forall k: \alpha_k=1$ is updated using pseudo-counts $N_k$, which are predicted by a neural network. This is different from prior networks, where the prior concentration parameters $\balpha$ are predicted instead. In both cases, this does not require any modification to a model's architecture except for replacing the softmax output function by a ReLU (or similar). \citet{sensoy2018evidential} for instance
train their model using the same techniques presented in the previous section: The main objective is the $l_2$ loss, penalizing the difference between the predicted Dirichlet and the one-hot encoded class label (\cref{app:il2-norm-loss}), and the KL divergence from a uniform  Dirichlet is used for regularization.


\paragraph{Generating OOD samples using generative models} Since OOD examples are not always readily available, several works try to create artificial samples using deep generative models.  \citet{hu2021multidimensional} train a Wasserstein GAN \citep{arjovsky2017wasserstein} to generate OOD samples, on which the network's uncertainty is maximized. The uncertainty is given through \emph{vacuity}, defined as $K / \sum_k \beta_k$. The vacuity compares a uniform prior belief against the amassed evidence $\sum_k \beta_k$, and thus is $1$ when there is no additonal evidence available. In a follow-up work, \citet{sensoy2020uncertainty} similarly train a model using a contrastive loss with artificial OOD samples from a Variational Autoencoder \citep{kingma2014autoencoding}, and a KL-based regularizer similar to that of \citet{tsiligkaridis2019information}, where the density for posterior concentration parameters $\beta_k$ that do not correspond to the true label are pushed to the uniform distribution.

\begin{figure}[tb]
    \begin{subfigure}[t]{0.41\linewidth}
        \centering
        \includegraphics[width=\textwidth]{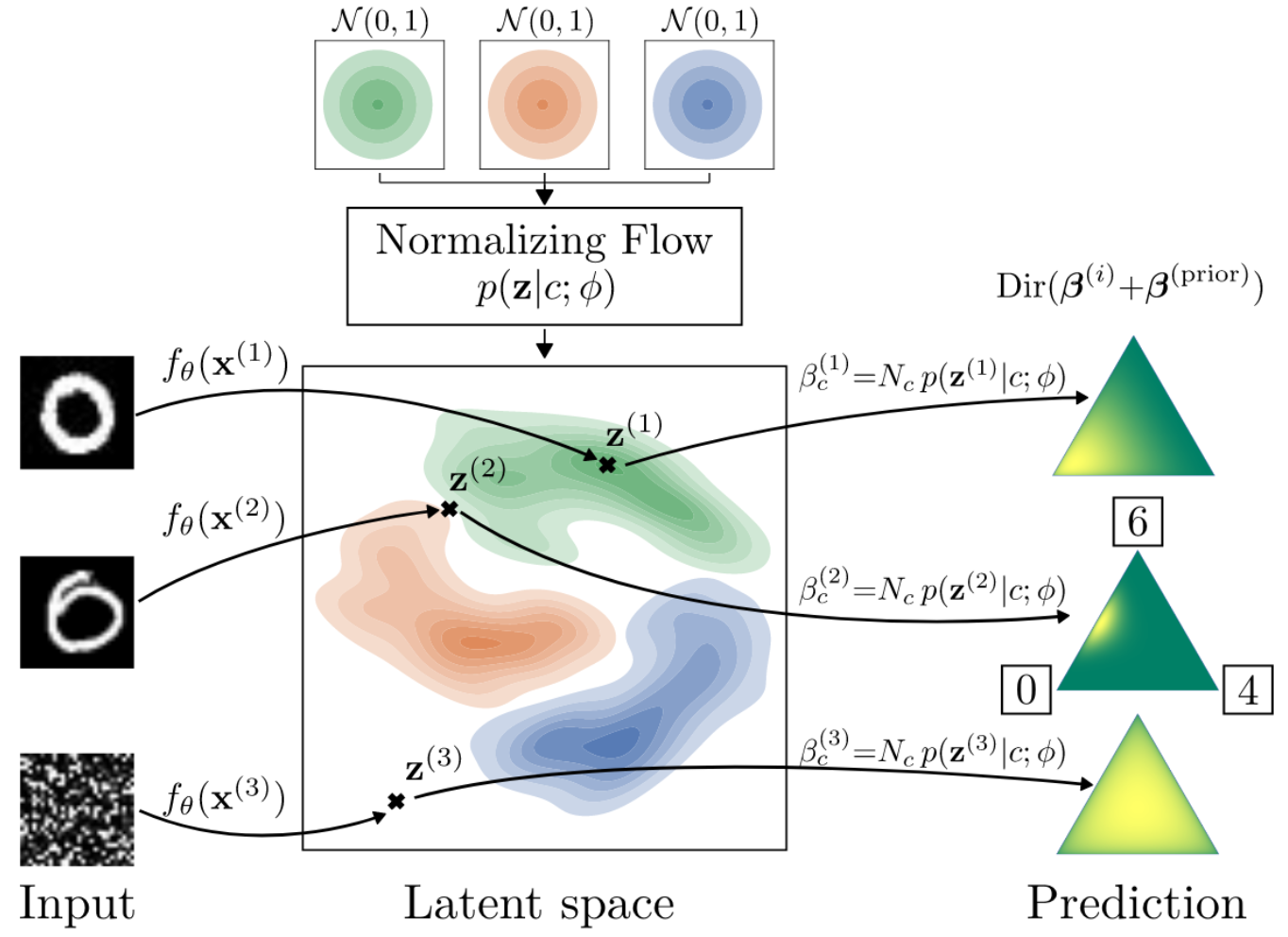}
        \caption{Posterior Network \citep{charpentier2020posterior}.}
        \label{subfig:posterior-network}
    \end{subfigure}
    \hfill
    \begin{subfigure}[t]{0.58\linewidth}
        \centering
        \includegraphics[width=\textwidth]{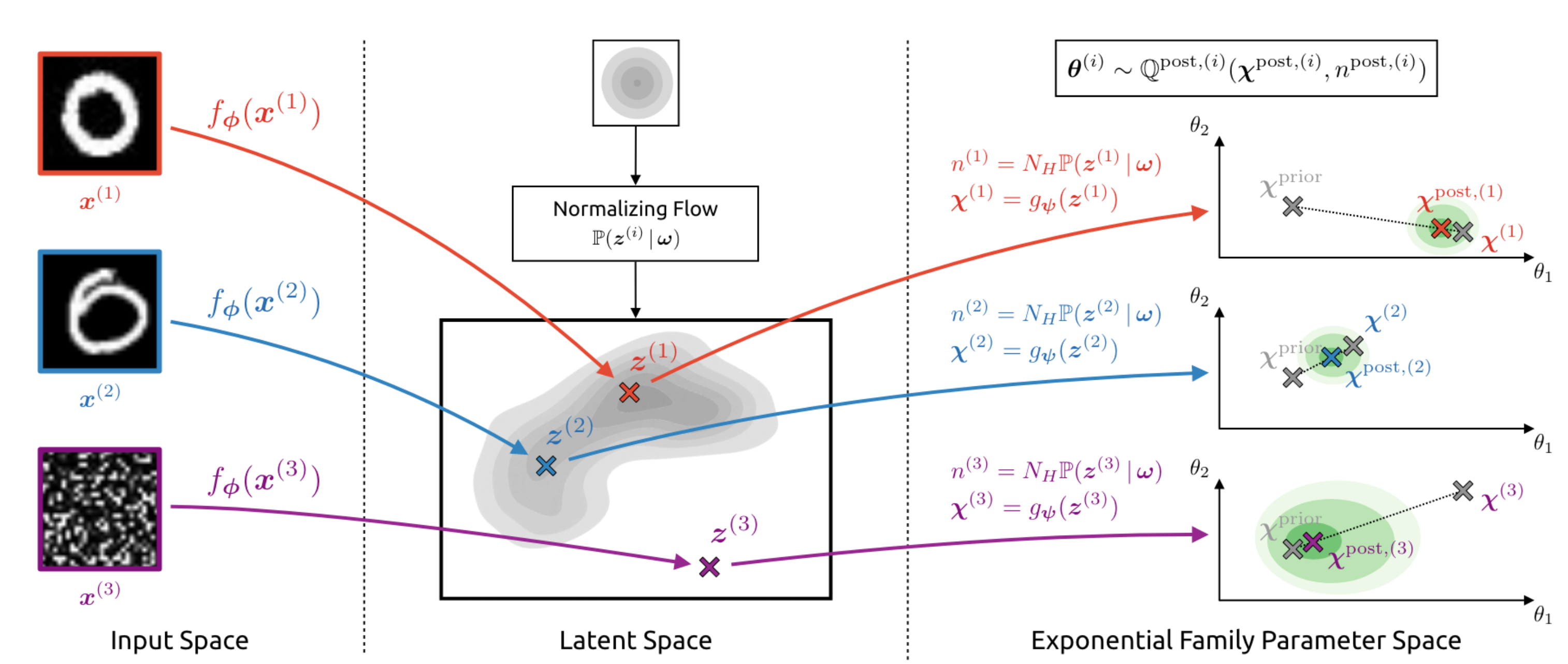}
        \caption{Natural Posterior Network \citep{charpentier2021natural}.}
        \label{subfig:natural-posterior-network}
    \end{subfigure}
    \caption{Schematic of the  Posterior Network and Natural Posterior Network, taken from \citet{charpentier2020posterior,charpentier2021natural}, respectively. In both cases, an encoder $f_{\bm{\theta}}$ maps inputs to a latent representation $\bz$. NFs then model the latent densities, which are used together with the prior concentration to produce the posterior parameters. In (a), the latent representation of $\bx^{(1)}$ lies right in the modelled density of the first class, and thus receives a confident prediction. The latent $\bz^{(2)}$ lies between densities, creating aleatoric uncertainty. $\bx^{(3)}$ is an OOD input, is mapped to a low-density area of the latent space and thus produces an uncertain prediction. The differences in the two approaches is that the Posterior Network in (a) uses one NF per class, while only one NF is used in (b). Furthermore, (b) constitutes a generalization to different exponential family distributions, and is not restricted to classification problems (see main text for more detail).}
    \label{fig:posterior-networks}
\end{figure}

\paragraph{Posterior networks via Normalizing Flows} \citet{charpentier2020posterior} also set $\bm{\alpha}$ to a uniform prior, but obtain the pseudo-observations $N_k$ in a different way: Instead of a model predicting them directly, $N_k$ is determined by the number of examples of a certain class in the training set. This quantity is further modified in the following way: An encoder model $f_{\bm{\theta}}$ produces a latent representation $\bz$ of some input. A (class-specific) normalizing flow\footnote{A NF is a generative model, estimating a density in the feature space by mapping it to a Gaussian in a latent space by a series of invertible, bijective transformations. The probability of an input can then be estimated by calculating the probability of its latent encoding under that Gaussian and applying the change-of-variable formula, traversing the flow in reverse. Instead of mapping from the feature space into latent space, the flows in \citet{charpentier2020posterior} map from the encoder latent space into a separate, second latent space.} (NF; \citeauthor{rezende2015variational}, \citeyear{rezende2015variational}) with parameters $\bphi$ then assigns a probability to this latent representation, which is used to weight $N_k$: 


\begin{equation}\label{eq:post-net-concentration}
    \beta_k = \alpha_k + N_k \cdot p(\bz | y=k, \bm{\phi});\quad \bz = f_{\bm{\theta}}(\bx).
\end{equation}

This has the advantage of producing low probabilities for strange inputs like the noise as depicted in \cref{subfig:posterior-network}, which in turn translate to low concentration parameters of the posterior Dirichlet, as it falls back onto the uniform prior. The model is optimized using the same uncertainty-aware cross-entropy loss as in \citet{bilovs2019uncertainty} with an additional entropy regularizer, encouraging density only around the correct class. This scheme is also applied to Graph Neural Networks by \citet{stadler2021graph}: In order to take the neighborhood structure of the graph into account, the authors also use a Personalized Page Rank scheme to diffuse node-specific posterior parameters $\bbeta$ between neighboring nodes. The Page Rank scores, reflecting the importance of a neighboring node to the current node, can be approximated using power iteration \citep{klicpera2019predict} and used to aggregate the originally predicted concentration parameters $\bbeta$ on a per-node basis. 

A generalization of the posterior network method to exponential family distributions is given by \citet{charpentier2021natural}. Akin to the update for the posterior Dirichlet parameters, the authors formulate a general Bayesian update rule as 

\begin{equation}\label{eq:natural-posterior-update}
    \bchi_i^\text{post} = \frac{n^\text{prior}\bchi^\text{prior} + n_i\bchi_i}{n^\text{prior} + n_i};\quad \bz_i = f_{\bm{\theta}}(\bx_i);\quad n_i = N \cdot p(\bz | \bm{\phi});\quad \bchi_i = g_{\bpsi}(\bx_i).
\end{equation}

$\bchi$ here denotes the parameters of the exponential family distribution and $n$ the evidence. Thus, posterior parameters for a sample $\bx_i$ are obtained by updating the prior parameter and some prior evidence by some input-dependent pseudo-evidence $n_i$ and parameters $\bchi_i$: Again, given a latent representation by an encoder $\bz$, a (this time single) normalizing flow predicts $n_i = N_H \cdot p(\bz | \bm{\phi})$ based on some pre-defined certainty budget $N_H$.\footnote{The certainty budget can simply be set to the number of available datapoints, however \citet{charpentier2021natural} suggest to set it to $\log N_H = \frac{1}{2}\big( H \log (2\pi) + \log(H+1)\big)$ to better scale with the dimensionality $H$ of the latent space.} The update parameters $\bchi_i$ are predicted by an additional network $\bchi_i = g_{\bpsi}(\bz)$, see \cref{subfig:natural-posterior-network}. For classification, $n^\text{prior} = 1$ and $\bchi^\text{prior}$ corresponds to the uniform Dirichlet, while $\bchi_i$ are concentration parameters predicted by an output layer based on the latent encoding. For unfamiliar inputs, this method will again result in a small pseudo-evidence term $n_i$, reflecting high model uncertainty. Since the generalization to the exponential family implies the application of this scheme to normal distributions, we will discuss the same method applied to regression in the next section. 

\paragraph{Posterior networks via variational inference} Another route lies in directly parameterizing the posterior parameters $\bm{\beta}$. Given a target distribution defined by a uniform Dirichlet prior plus the number of times an input is associated with a specific label, \citet{chen2018variational} optimize a distribution matching objective, i.e.\@ the KL-divergence between the posterior parameters predicted by a neural network and the target distribution. Since this objective is intractable to optimize directly, this leaves us to instead model an \emph{approximate posterior} using variational inference methods. As the KL divergence between the true and approximate posterior is infeasible to estimate, variational methods usually optimize the \emph{evidence lower bound} (ELBO):


\begin{equation}\label{eq:elbo}
    \mathcal{L}_\text{ELBO} = \underbrace{\psi(\beta_y) - \psi(\beta_0)}_{\text{UCE loss}} - \underbrace{\log\frac{B(\bm{\beta})}{B(\bm{\gamma})}+ \sum_{k=1}^K (\beta_k - \gamma_k)\big(\psi(\beta_k) - \psi(\beta_0)\big)}_{\text{KL-divergence}}
\end{equation}

in which we can identify to consist of the uncertainty-aware cross-entropy (UCE) loss used by \citet{bilovs2019uncertainty, charpentier2020posterior, charpentier2021natural} and the KL-divergence between two Dirichlets (\cref{app:kl-dirichlets}). This approach is also employed by \citet{joo2020being}, \citet{chen2018variational} and \citet{shen2022post}, while the latter predict posterior parameters based on the activations of different layers of a pre-trained feature extractor.


\section{Evidential Deep Learning for Regression}\label{sec:evidential-regression}

\begin{table}[tb]
    \centering
    \caption{Overview over Evidential Deep Learning methods for regression.}\label{table:overview-regression}
    
     \resizebox{0.95\textwidth}{!}{
        \renewcommand{\arraystretch}{1.6}
        \begin{tabular}{@{}llll@{}}
            \toprule
            Method & \thead[tl]{Parameterized \\distribution} & Loss function & Model \\
            \midrule
            \makecell[tl]{Deep Evidential Regression \\
            \citep{amini2020deep}} & \makecell[tl]{Normal-Inverse\\ Gamma Prior} & \makecell[tl]{Negative log-likelihood loss + KL\\ w.r.t. uniform prior } & MLP / CNN \\ 
            \makecell[tl]{Deep Evidential Regression\\with Multi-task Learning\\ \citep{oh2021improving}} & \makecell[tl]{Normal-Inverse\\ Gamma Prior} & \makecell[tl]{Like \citet{amini2020deep}, with additional\\ Lipschitz-modified MSE loss} & MLP / CNN \\
            \makecell[tl]{ Multivariate Deep Evidential \\ Regression \citep{meinert2021multivariate}} & \makecell[tl]{Normal-Inverse\\Wishart Prior} & \makecell[tl]{Like \citet{amini2020deep}, but tying two\\ predicted params. instead of using a regularizer} & MLP \\
            \makecell[tl]{Regression Prior Network\\ \citep{malinin2020regression}} & Normal-Wishart Prior & \makecell[tl]{Reverse KL \\ \citep{malinin2019reverse}} & MLP / CNN \\
            \makecell[tl]{Natural Posterior Network \\
            \citep{charpentier2021natural}} & \makecell[tl]{Inverse-$\chi^2$ Posterior} & \makecell[tl]{Uncertainty Cross-entropy \citep{bilovs2019uncertainty} \\ + Entropy regularizer} & \makecell[tl]{MLP / CNN +\\Norm. Flow } \\
            \bottomrule
        \end{tabular}%
    }
\end{table}

Because the EDL framework provides convenient uncertainty estimation, the question naturally arises of whether it can be extended to regression problems as well. The answer is affirmative, although the Dirichlet distribution is not an appropriate choice in this case. It is very common to model a regression problem using a normal likelihood (\citeauthor{bishop2006pattern}, \citeyear{bishop2006pattern}; Chapter 3.3). As such, there are multiple potential choices for a prior distribution. The methods listed in \cref{table:overview-regression} either choose the Normal-Inverse Gamma distribution \citep{amini2020deep, charpentier2021natural}, inducing a scaled inverse-$\chi^2$ posterior \citep{gelman1995bayesian},\footnote{The form of the Normal-Inverse Gamma posterior and the Normal Inverse-$\chi^2$ posterior are interchangable using some parameter substitutions \citep{murphy2007conjugate}.} or a Normal-Wishart prior \citep{malinin2020regression}. We will discuss these approaches in turn.

\begin{figure}[tb]
    \centering 
    \includegraphics[width=0.9\textwidth]{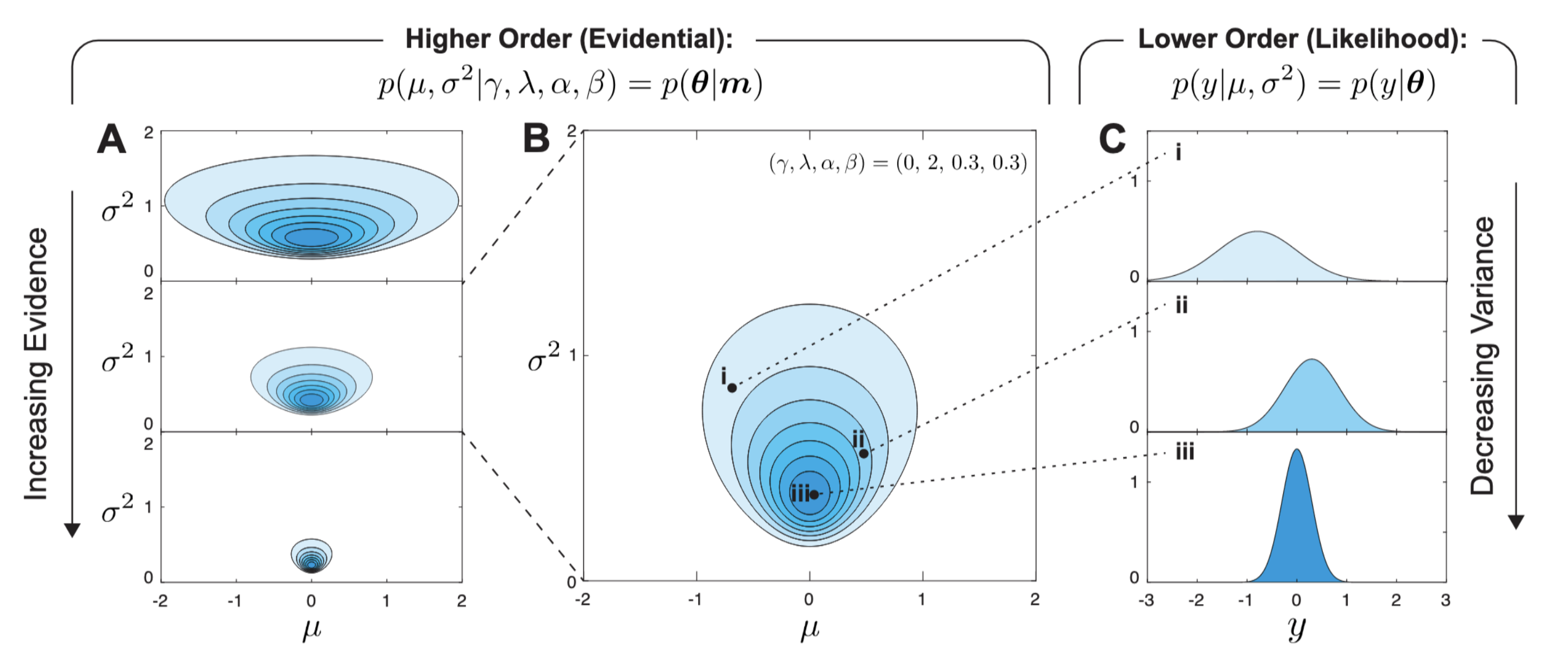}
    \caption{Example of an application of Evidential Deep Learning for regression, taken from \citet{amini2020deep}. The neural network predicts an Normal Inverse-Gamma prior, whose corresponding normal likelihoods display decreasing variance (and thus uncertainty) in the face of stronger evidence. }\label{fig:evidential-regression} 
\end{figure}

\paragraph{Univariate regression} \citet{amini2020deep} model the regression problem as a normal distribution with unknown mean and variance $\mathcal{N}(y; \pi, \sigma^2)$, and use a normal prior for the mean with $\pi \sim \mathcal{N}(\gamma, \sigma^2\nu^{-1})$ and an inverse Gamma prior for the variance with $\sigma^2 \sim \Gamma^{-1}(\alpha, \beta)$, resulting in a combined Inverse-Gamma prior with parameters $\gamma, \nu, \alpha, \beta$, shown in \cref{fig:evidential-regression}. These are predicted by different ``heads'' of a neural network. For predictions, the expectation of the mean corresponds to $\Expect[]{}{\pi} = \gamma$, and aleatoric and epistemic  uncertainty can then be estimated using the expected value of the variance as well as the variance of the mean, respectively, which have closed form solutions under this parameterization:

\begin{equation}
    \Expect[]{}{\sigma^2} = \frac{\beta}{\alpha - 1};\quad \text{Var}[\pi] = \frac{\beta}{\nu(\alpha - 1)}
\end{equation}

By choosing to optimize using a negative log-likelihood objective, 
we can actually evaluate the loss function analytically, since the likelihood function corresponds to a Student's t-distribution with $\gamma$ degrees of freedom, mean $\beta(1 + \nu)/(\nu\alpha)$ and $2\alpha$ variance:

\begin{equation}\label{eq:evidential-regression-loglik}
    \mathcal{L}_\text{NLL} = \frac{1}{2}\log\Big(\frac{\pi}{\nu}\Big) - \alpha\log \Omega + \Big(\alpha + \frac{1}{2}\Big)\log \Big((y_i - \gamma)^2\nu + \Omega \Big) + \log\bigg(\frac{\Gamma(\alpha)}{\Gamma(\alpha + \frac{1}{2})}\bigg)
\end{equation}

using $\Omega = 2\beta(1 + \nu)$. Akin to the entropy regularizer for Dirichlet networks, \citet{amini2020deep} propose a regularization term that only allows for concentrating density on the correct prediction:
 
 \begin{equation}\label{eq:evidential-regression-regularizer}
     \mathcal{L}_\text{reg} = |y_i - \gamma | \cdot (2\nu + \alpha)
 \end{equation}

Since $\Expect[]{}{\pi} = \gamma$ is the prediction of the network, the second term in the product will be scaled by the degree to which the current prediction deviates from the target value. Since $\nu$ and $\alpha$ control the variance of the mean and the variance of the normal likelihood, this term encourages the network to decrease the evidence for mispredicted data samples. As \citet{amini2020deep} point out, large amounts of evidence are not punished in cases where the prediction is close to the target.
However, \citet{oh2021improving} argue that this combination of objectives might create adverse incentives for the model during training: Since the difference between the prediction and target in \cref{eq:evidential-regression-loglik} is scaled by $\nu$, the model could learn to increase the predictive uncertainty by decreasing $\nu$ instead of improving its prediction. They propose to ameliorate this issue by using a third loss term of the form 

\begin{equation}\label{eq:evidential-regression-mse-loss}
    \mathcal{L}_\text{MSE} = \begin{cases}
    (y_i - \gamma)^2       & \quad \text{if } (y_i - \gamma)^2 < U_{\nu, \alpha}\\
    2 \sqrt{U_{\nu, \alpha}}|y_i - \gamma| - U_{\nu, \alpha}  & \quad \text{if } (y_i - \gamma)^2 \ge U_{\nu, \alpha}
  \end{cases}
\end{equation}

where $U_{\nu, \alpha} = \min(U_\nu, U_\alpha)$ denotes the minimum value for the uncertainty thresholds for $\nu, \alpha$ given over a mini-batch, which are themselves defined as 

\begin{equation}\label{eq:evidential-regression-mse-loss-bounds}
    U_\nu = \frac{\beta(\nu + 1)}{\alpha\nu};\quad 
    U_\alpha = \frac{2\beta(\nu + 1)}{\nu}\Big[\exp\Big(\psi\Big(\alpha + \frac{1}{2}\Big) - \psi(\alpha)) - 1\Big)\Big].\\
\end{equation}

These expression are obtained by taking the derivatives $\partial \mathcal{L}_\text{NLL}/\partial \nu$, $\partial \mathcal{L}_\text{NLL}/\partial \alpha$ and solving for the parameters, thus giving us the values for $\nu$ and $\alpha$ for which the loss gradients are maximal. In combination with \cref{eq:evidential-regression-mse-loss}, \cref{eq:evidential-regression-mse-loss-bounds} ensures that, should the model error exceed $U_{\nu, \alpha}$, the error is rescaled. Thus, this rescaling bounds the Lipschitz constant of the loss function, motivating the model to ensure the correctness of its prediction, since its ability to increase uncertainty to decrease its loss is now limited. 

\paragraph{Posterior networks for regression} Another approach for regression is the Natural Posterior Network by \citet{charpentier2021natural}, which was already discussed for classification in \cref{sec:posterior-networks}. But since the proposed approach is a generalization for exponential family distributions, it can be applied to regression as well, using a Normal likelihood and Normal Inverse-Gamma prior. The Bayesian update rule in \cref{eq:natural-posterior-update} is adapted as follows: $n$ is set to $n = \lambda = 2\alpha$, and $\bchi = \big[\pi_0\ |\ \pi_0^2 + 2\beta / n\big]^T$. Feeding an input into the natural posterior network again first produces a latent encoding $\bz$, from which a NF predicts $n_i = N_H \cdot p(\bz | \bm{\phi})$, and an additional network produces $\bchi_i = g_{\bpsi}(\bz)$, which are used in \cref{eq:natural-posterior-update} to produce $\bchi^\text{post}$ and $n^\text{post}$, from which the parameters of the posterior Normal Inverse-Gamma can be derived. The authors also produce a general exponential family form of the UCE loss by \citet{bilovs2019uncertainty}, consisting of expected log-likelihood and an entropy regularizer, which they derive for the regression parameterization. Again, this approach relies on the density estimation capabilities of the NF to produce an agnostic belief about the right prediction for OOD examples (see \cref{subfig:natural-posterior-network}).


\paragraph{Multivariate evidential regression} There are also some works offering solutions for multivariate regression problems:  \citet{malinin2020regression} can be seen as a multivariate generalization of the work of \citet{amini2020deep}, where a combined Normal-Wishart prior is formed to fit the now Multivariate Normal likelihood. Again, the prior parameters are the output of a neural network, and uncertainty can be quantified in a similar way. For training purposes, they apply two different training objectives using the equivalent of the reverse KL objective of \citet{malinin2019reverse} as well as of the knowledge distillation objective of \citet{malinin2020ensemble}, which does not require OOD data for regularization purposes. \citet{meinert2021multivariate} also provide a solution using a Normal Inverse-Wishart prior. In a similar vein to \citet{oh2021improving}, they argue that the original objective proposed by \citet{amini2020deep} can be minimized by increasing the network's uncertainty instead of decreasing the mismatch of its prediction. As a solution, they simply propose to tie $\beta$ and $\nu$ via a hyperparameter.

\section{Related Work}\label{sec:related-work}

\paragraph{Other Approaches to Uncertainty Quantification} The need for the quantification of uncertainty in order to earn the trust of end-users and stakeholders has been a key driver for research \citep{bhatt2021uncertainty, jacovi2021formalizing, liao2022designing}. Existing methods can broadly be divided into frequentist and Bayesian methods, where the former judge the confidence of a model based on its predicted probabilities. Unfortunately, standard neural discriminator architectures have been proven to possess unwanted theoretical properties w.r.t.\@ OOD inputs  \citep{hein2019why, ulmer2021know} and might therefore be unable to detect potentially risky inputs.\footnote{\citet{pearce2021understanding} argue that some insights might partially be misled by low-dimensional intuitions, and that empirically OOD data in higher dimensions tend to be mapped into regions of higher uncertainty.} Further, a large line of research works has questioned the calibration of models \citep{guo2017on, nixon2019measuring, desai2020calibration, minderer2021revisiting, wang2021rethinking}, i.e.\@ to what extend the probability score of a class---also referred to as its confidence---corresponds to the chance of a correct prediction. Instead of relying on the confidence score alone, another way lies in constructing prediction sets consisting of the classes accumulating a certain share of the total predictive mass \citep{kompa2021empirical, ulmer2022exploring}. By scoring a held-out population of data points to calibrate these prediction sets, we can also obtain frequentist guarantees in a procedure referred to a \emph{conformal prediction} \citep{papadopoulos2002inductive, vovk2005algorithmic, lei2014distribution, angelopoulos2021gentle}. This however still does not let us distinguish different notions of uncertainty. 
A popular \emph{Bayesian} way to overcome this blemish by aggregating multiple predictions by networks in the Bayesian model averaging framework \citep{mackay1992bayesian, mackay1995developments, hinton1993keeping, neal2012bayesian, jeffreys1998theory, wilson2020bayesian, kristiadi2020being, daxberger2021laplace, gal2016dropout, blundell2015weight, lakshminarayanan2017simple}. Nevertheless, many of these methods have been shown not to produce diverse predictions \citep{wilson2020bayesian, fort2019deep} and to deliver subpar performance and potentially misleading uncertainty estimates under distributional shift \citep{ovadia2019can, masegosa2019learning, wenzel2020good, izmailov2021dangers, izmailov2021what}, raising doubts about their efficacy.
The most robust method in this context is often given by an ensemble of neural predictors \citep{lakshminarayanan2017simple}, with multiple works exploring ways to make their training more efficient \citep{huang2017snapshot, wilson2020bayesian, wen2020batchensemble, turkoglu2022film} or to provide theoretical guarantees \citep{pearce2020uncertainty, ciosek2020conservative, he2020bayesian, d2021repulsive}.  

\paragraph{Related Approaches to EDL} \citet{kull2019beyond} found an appealing use of the Dirichlet distribution as a post-training calibration map. \citet{hobbhahn2022fast} use the Laplace bridge, a modified inverse based on an idea by \citet{mackay1998choice}, to map from the model's logit space to a Dirichlet distribution. The proposed Posterior Network \citep{charpentier2020posterior, charpentier2021natural} can furthermore be seen as related to another, competing approach, namely the combination of neural discriminators with density estimation methods, for instance in the form of energy-based models \citep{grathwol2020your, elflein2021out} or other hybrid architectures \citep{lee2018simple, mukhoti2021deterministic}. Furthermore, there is a line of other single-pass uncertainty quantification approaches which do not originate from the evidential framework, for instance by taking inspiration from RBF networks \citep{van2020uncertainty} or via Gaussian Process output layers \citep{liu2020simple, fortuin2021deep, van2021feature}. 

\paragraph{Applications of EDL} Some of the discussed models have already found a variety of applications, such as in autonomous driving \citep{capellier2019evidential, liu2021efficient, petek2021robust, wang2021uncertainty}, remote sensing \citep{gawlikowski2022advanced}, medical screening \citep{ghesu2019quantifying, gu2021efficient, li2022region}, molecular analysis \citep{soleimany2021evidential}, open set recognition \citep{bao2021evidential}, active learning \citep{hemmer2022deal} and model selection \citep{radev2021amortized}.

\section{Discussion}\label{sec:discussion}

\paragraph{What is state-of-the-art?} As apparent from \cref{table:overview-evaluation}, evaluation methods and datasets can vary tremendously between different research works (for an overview, refer to  \cref{app:dataset-evaluation},). This can make it hard to accurately compare different approaches in a fair manner. Nevertheless, we try to draw some conclusion about the state-of-art in this research direction to the best extent possible: For \textbf{image classification}, the posterior \citep{charpentier2020posterior} and natural posterior network \citep{charpentier2021natural} provide the best results on the tested benchmarks, both in terms of task performance and uncertainty quality. When the training an extra normalizing flow creates too much computational overhead, prior networks \citep{malinin2018predictive} with the PAC-based regularizer (\citeauthor{haussmann2019bayesian}, \citeyear{haussmann2019bayesian}; see \cref{tab:overview-loss} for final form) or a simple entropy regularizer (\cref{app:entropy-dirichlet}) can be used. In the case of \textbf{regression} problems, the natural posterior network \citep{stadler2021graph} performs better or on par with the evidential regression by \citet{amini2020deep} or an  ensemble \citet{lakshminarayanan2017simple} or MC Dropout \citep{gal2016dropout}. For \textbf{graph neural networks}, the graph posterior network \citep{stadler2021graph} and a ensemble provide similar performance, but with the former displaying better uncertainty results. Again, this model requires training a NF, so a simpler fallback is provided by evidential regression \citep{amini2020deep} with the improvement by \citet{oh2021improving}. For \textbf{NLP} and \textbf{count prediction}, the works of \citet{shen2020modeling} and \citet{charpentier2021natural} are the only available instances from this model family, respectively. In the latter case, ensembles and the evidential regression framework \citep{amini2020deep} produce a lower root mean-squared error, but worse uncertainty estimates on OOD.\\

\paragraph{Computational Cost} When it comes to the computational requirements, most of the proposed methods in this survey incur the same cost as a single deterministic network using a softmax output, since most of the architecture remains unchanged. Additional cost is mostly only produced when using knowledge distillation \citep{malinin2020ensemble, fathullah2022self}, adding normalizing flow components like for posterior networks \citep{charpentier2020posterior, charpentier2021natural, stadler2021graph} or using generative models to produce synthetic OOD data \citep{chen2018variational, sensoy2020uncertainty, hu2021multidimensional}.

\paragraph{Comparison to Other Approaches to Uncertainty Quantification} As discussed in \cref{sec:related-work}, several existing approaches to uncertainty quantification equally suffer from shortcomings with respect to their reliability. One possible explanation for this behavior might lie in the insight that neural networks trained in the empirical risk minimization framework tend to learn spurious but highly predictive features \citep{ilyas2019adversarial, nagarajan2020understanding}. This way, inputs stemming from the training distribution can be mapped to similar parts of the latent space as data points outside the distribution even though they display (from a human perspective) blatant semantic differences, simply because these semantic features were not useful to optimize for the training objective. This can result in ID and OOD points having assigned similar feature representations by a network, a phenomenon has been coined ``feature collapse'' \citep{nalisnick2019do,van2021feature,havtorn2021hierarchical}. One strategy to mitigate (but not solve) this issue has been to enforce a constraint on the smoothness of the neural network function \citep{wei2018improving, amersfoort2020uncertainty, van2021feature, liu2020simple}, thereby maintaining both a sensitivity to semantic changes in the input and robustness against adversarial inputs \citep{yu2019interpreting}. Another approach lies in the usage of OOD data as well, sometimes dubbed ``outlier exposure'' \citep{fort2021exploring}, but displaying the same shortcomings as in the EDL case. A generally promising strategy seems to seek functional diversity through ensembling: \citet{juneja2022linear} show how model instances ending up in different low-loss modes correspond to distinct generalization strategies, indicating that combining diverse strategies may lead to better generalization and thus potentially more reliable uncertainty. Attaining different solutions still creates computational overhead, despite new methods to reduce it \citep{garipov2018loss, dusenberry2020efficient, benton2021loss}. 

\paragraph{Bayesian model averaging} One of the most fundamental differences between EDL and existing approaches is the sacrifice of Bayesian model averaging (\cref{eq:bma,eq:prior-networks-fac}): In principle, combining multiple parameter estimates is supposed to result in a lower predictive risk \citep{fragoso2018bayesian}. The Machine Learning community has ascribed further desiderata to this approach, such as better generalization and robustness to distributional shifts. Recent studies with exact Bayesian Neural Networks however have cast doubts on these assumptions \citep{izmailov2021dangers, izmailov2021what}. Nevertheless, ensembles, that approximate \cref{eq:bma} via Monte Carlo estimates, remain state-of-the-art on many uncertainty benchmarks. EDL abandons modelling epistemic uncertainty through the learnable parameters, and instead expresses it through the uncertainty in prior / posterior parameters. This loses functional diversity which could aid generalization, while sidestepping computational costs. Future research could therefore explore the combination of both paradigms, as proposed by \citet{haussmann2019bayesian,zhao2020uncertainty,charpentier2021natural}.

\paragraph{Challenges} Despite their advantages, the last chapters have pointed out key weaknesses of Dirichlet networks as well: In order to achieve the right behavior of the distribution and thus guarantee sensible uncertainty estimates (since ground truth estimates are not available), the surveyed literature proposes a variety of loss functions. \citet{bengs2022difficulty} show formally that many of the loss functions used so far are \emph{not} appropriate and violate basic asymptotic assumptions about epistemic uncertainty: With increasing amount of data, epistemic uncertainty should vanish, but this is not guaranteed using the commonly used loss functions. Furthermore, some approaches \citep{malinin2018predictive, malinin2019reverse, nandy2020towards, malinin2020regression} require out-of-distribution data points during training. This comes with two problems: Such data is often not available or in the first place, or cannot guarantee robustness against \emph{other} kinds of unseen OOD data, of which infinite types exist in a real-valued feature space.\footnote{The same applies to the artificial OOD data in \citet{chen2018variational, shen2020modeling, sensoy2020uncertainty}.} Indeed, \citet{kopetzki2021evaluating} found OOD detection to deteriorate across a family of EDL models under adversarial perturbation and OOD data.  \citet{stadler2021graph} point out that much of the ability of posterior networks stems from the addition of a NF, which have been shown to also sometimes behave unreliably on OOD data \citep{nalisnick2019do}. Although the NFs in posterior networks operate on the latent and not the feature space, they are also restricted to operate on features that the underlying network has learned to recognize. Recent work by \citet{dietterich2022familiarity} has hinted at the fact that networks might identify OOD by the absence of known features, and not by the presence of new ones, providing a case in which posterior networks are likely to fail. Such evidence on OOD data and adversarial examples has indeed been identified by a study by \citet{kopetzki2021evaluating}. 

\paragraph{Future Research Directions} Overall, the following directions for future research on EDL crystallize from our previous reflections: \emph{(1) Explicit epistemic uncertainty estimation:} Since we often employ the point estimate in \cref{eq:prior-networks-fac} to avoid the posterior $p(\btheta|\mathbb{D})$, explicit estimation of the epistemic uncertainty is not possible, and some summary statistic of the concentration parameters is used for classification problems instead (\cref{sec:uncertainty-dirichlet}). Estimating model uncertainty through modelling the (approximate) posterior $p(\btheta|\mathbb{D})$ in Bayesian model averaging is a popular technique \citep{houlsby2011bayesian, gal2016uncertainty, gal2018understanding, ulmer2020trust}, but comes with the disadvantage of additional computational overhead. However, \citet{sharma2022bayesian} recently showed that a Bayesian treatment of all model parameters may not be necessary, potentially allowing for a compromise. \emph{(2) Robustness to diverse OOD data:} The emprical evidence compiled by \citet{kopetzki2021evaluating} indicates that EDL classification models are not completely able to robustly classify and detect OOD and adversarial inputs. These findings hold both for prior networks trained with OOD data, or for posterior networks using density estimators. We speculate that through the information bottleneck principle \citep{tishby2015deep}, EDL models might not learn input features that are useful to indicate uncertainty in their prediction, or at best identify the absence of known features, but not the presence of new ones \citep{dietterich2022familiarity}. Finding a way to have models identify unusual features could this help to mitigate this problem. \emph{(3) Theoretical guarantees:} Even though some guarantees have been derived for EDL classifiers w.r.t. OOD data points \citep{charpentier2020posterior, stadler2021graph}, \citet{bengs2022difficulty} point out the flaws of current training regimes for epistemic uncertainty in the limit of infinite limit. Furthermore, \citet{huellermeier2021aleatoric} argue that even uncertainty estimates are affected by uncertainty themselves, impacting their usefulness.

\section{Conclusion}

This survey has given an overview over contemporary approaches for uncertainty estimation using neural networks to parameterize conjugate priors or the corresponding posteriors instead of likelihoods, called Evidential Deep Learning. We highlighted their appealing theoretical properties allowing for uncertainty estimation with minimal computational overhead, rendering them as a viable alternative to existing strategies. We also emphasized practical problems: In order to nudge models towards the desired behavior in the face of unseen or out-of-distribution samples, the design of the model architecture and loss function have to be carefully considered. Based on a summary and discussion of experimental findings in \cref{sec:discussion}, the entropy regularizer seems to be a sensible choice in prior networks when OOD data is not available. Combining discriminators with generative models like normalizing flows as in \citet{charpentier2020posterior, charpentier2021natural}, embedded in a sturdy Bayesian framework, also appears as an exciting direction for practical applications. In summary, we believe that recent advances show promising results for Evidential Deep Learning, making it a viable option in uncertainty estimation  to improve safety and trustworthiness in Machine Learning systems. 

\section*{Acknowledgements}

We would like to thank Giovanni Cin\`a, Max M\"uller-Eberstein, Daniel Varab and Mike Zhang for reading early versions of this draft and providing tremendously useful feedback. Further, we would like to explicitly thank Mike Zhang for helping to improve \cref{fig:taxonomy}. We also would like to thank Alexander Amini for providing a long list of references that helped to further improve the coverage of this work and the anonymous reviewers for their suggestions. Lastly, we owe our gratitude to the anonymous reviewers that helped us such much to improve the different versions of this paper. 

\bibliography{tmlr}
\bibliographystyle{tmlr}

\clearpage

\appendix 

\section{Code Appendix}\label{app:code-appendix}

\subsection{Iris Example Training Details}\label{app:iris-code-details}

The code used to produce \cref{fig:iris-example} is available online.\footnote{Code will be made available upon acceptance.} All models use three layers with $100$ hidden units and ReLU activations each. We furthermore optimized all of the models with a learning rate of $0.001$ using the Adam optimizer \citep{kingma2014adam} with its default parameter settings. We also  regularize the ensemble and MC Dropout model with a dropout probability of $0.1$ each.

\paragraph{Prior Network specifics} We choose the expected $l_2$ loss by \citet{sensoy2018evidential} and regularize the network using the KL divergence w.r.t. to a uniform Dirichlet as in \citet{sensoy2018evidential}. In the regularization term, we do not use the original concentration parameters $\balpha$, but a version in which the concentration of the parameter $\alpha_k$ corresponding to the correct class is removed using a one-hot label encoding $\by$ by $\tilde{\balpha} = (1 - \balpha) \odot \balpha + \by \odot \balpha$, where $\odot$ denotes point-wise multiplication. The regularization term is added to the loss using a weighting factor of $0.05$.

\subsection{Code Availability}\label{app:code-availability}

\begin{table}[tb]
    \centering
    \caption{Overview over code repositories of surveyed works.}
     \resizebox{0.95\textwidth}{!}{
        \renewcommand{\arraystretch}{1.6}
        \begin{tabular}{@{}ll@{}}
            \toprule
            Paper & Code Repository \\
            \midrule
            \makecell[tl]{Prior network\\ \citep{malinin2018predictive}}       &  \url{https://github.com/KaosEngineer/PriorNetworks-OLD} \\
            \makecell[tl]{Prior networks\\ \citep{malinin2019reverse}}         &  \url{https://github.com/KaosEngineer/PriorNetworks} \\
            \makecell[tl]{Dirichlet via Function Decomposition\\ \citep{bilovs2019uncertainty}} & \url{https://github.com/sharpenb/Uncertainty-Event-Prediction} \\ 
            \makecell[tl]{Prior network with PAC Regularization \\ \citep{haussmann2019bayesian}} &  \url{https://github.com/manuelhaussmann/bedl} \\
            \makecell[tl]{Prior networks with representation gap\\ \citep{nandy2020towards}} & \url{https://github.com/jayjaynandy/maximize-representation-gap} \\
            \makecell[tl]{Graph-based Kernel Dirichlet distribution\\estimation (GKDE)
            \citep{zhao2020uncertainty}} & \url{https://github.com/zxj32/uncertainty-GNN} \\
            \makecell[tl]{Evidential Deep Learning\\  \citep{sensoy2018evidential}} & \url{https://muratsensoy.github.io/uncertainty.html} \\
            \makecell[tl]{WGAN--ENN\\ \citep{hu2021multidimensional}} & \url{https://github.com/snowood1/wenn} \\
            \makecell[tl]{Belief Matching  \citep{joo2020being}}           & \url{https://github.com/tjoo512/belief-matching-framework} \\ 
            \makecell[tl]{Posterior Networks\\ \citep{charpentier2020posterior}} & \url{https://github.com/sharpenb/Posterior-Network} \\
            \makecell[tl]{Graph Posterior Networks\\ \citep{stadler2021graph}} & \url{https://github.com/stadlmax/Graph-Posterior-Network}  \\
            \makecell[tl]{Generative Evidential Neural Networks\\ \citep{sensoy2020uncertainty}} & \url{https://muratsensoy.github.io/gen.html} \\ 
            \makecell[tl]{Deep Evidential Regression\\with Multi-task Learning\\ \citep{oh2021improving}} & \url{https://github.com/deargen/MT-ENet} \\
            \makecell[tl]{ Multivariate Deep Evidential \\ Regression \citep{meinert2021multivariate}} & \url{https://github.com/avitase/mder/} \\
            \makecell[tl]{Regression Prior Network\\ \citep{malinin2020regression}} &  \url{https://github.com/JanRocketMan/regression-prior-networks} \\
            \makecell[tl]{Natural Posterior Network \\
            \citep{charpentier2021natural}} & \url{https://github.com/borchero/natural-posterior-network} \\
            \bottomrule
        \end{tabular}%
    }
    \label{tab:code-availability}
\end{table}

We list the available code repositories for surveyed works in \cref{tab:code-availability}. Works for which no official implementation could be found are not listed.

\section{Datasets \& Evaluation Techniques Appendix}\label{app:dataset-evaluation}

\begin{table}[ht!]
    \centering
    \caption{Overview over uncertainty evaluation techniques and datasets. $^{(*)}$ indicates that a dataset was used as an OOD dataset for evaluation purposes, while $^{(\diamond)}$ signifies that it was used as an in-distribution or out-of-distribution dataset. $^{(\dagger)}$ means that a dataset was modified to create ID and OOD splits (for instance by removing some classes for evaluation or corrupting samples with noise). }\label{table:overview-evaluation}
    \resizebox{0.985\textwidth}{!}{
        \renewcommand{\arraystretch}{2.4}
        \begin{tabular}{@{}llccc@{}}
            \toprule
            & & \multicolumn{3}{c}{\textbf{Data Modality}} \\
            Method & \thead[tl]{Uncertainty Evaluation} & Images & Tabular & Other \\
            \midrule
            \makecell[tl]{Prior network\\ \citep{malinin2018predictive}} & \makecell[ml]{OOD Detection,\\ Misclassification Detection} & \makecell[ml]{MNIST, CIFAR-10,\\Omniglot$^{(*)}$, SVHN$^{(*)}$, \\LSUN$^{(*)}$, TIM$^{(*)}$} & \xmark & Clusters (Synthetic) \\
            \makecell[tl]{Prior networks\\ \citep{malinin2019reverse}} & \makecell[ml]{OOD Detection,\\ Adversarial Attack Detection} & \makecell[ml]{MNIST, CIFAR-10/100,\\ SVHN$^{(*)}$, LSUN$^{(*)}$, TIM$^{(*)}$} & \xmark & Clusters (Synthetic) \\
            \makecell[tl]{Information Robust Dirichlet Networks\\ \citep{tsiligkaridis2019information}} & \makecell[ml]{OOD Detection,\\ Adversarial Attack Detection} & \makecell[ml]{MNIST, FashionMNIST$^{(*)}$\\ notMNIST$^{(*)}$, Omniglot$^{(*)}$\\CIFAR-10, TIM$^{(*)}$, SVHN$^{(*)}$} & \xmark & \xmark  \\
            \makecell[tl]{Dirichlet via Function Decomposition\\ \citep{bilovs2019uncertainty}} & OOD Detection & \xmark & \makecell[ml]{Erd\H{o}s-R\'{e}nyi Graph\\(Synthetic), Stack Exchange,\\ Smart Home, Car Indicators} & \xmark \\ 
            \makecell[tl]{Prior network with PAC Regularization \\ \citep{haussmann2019bayesian}} & OOD Detection & \makecell[ml]{MNIST, FashionMNIST$^{(*)}$\\CIFAR-10$^{(\dagger)}$} & \xmark & \xmark  \\
            \makecell[tl]{Ensemble Distribution  Distillation\\ \citep{malinin2020ensemble}} & \makecell[ml]{OOD Detection,\\ Misclassification Detection,\\Calibration} & \makecell[ml]{CIFAR-10, CIFAR-100$^{(\diamond)}$\\TIM$^{(\diamond)}$, LSUN$^{(*)}$} & \xmark & Spirals (Synthetic) \\
            \makecell[tl]{Self-Distribution Distillation\\ \citep{fathullah2022self}} & \makecell[ml]{OOD Detection,\\Calibration} & \makecell[ml]{CIFAR-100\\SVHN$^{(*)}$, LSUN$^{(*)}$} & \xmark & \xmark \\
            
            \makecell[tl]{Prior networks with representation gap\\ \citep{nandy2020towards}} & OOD Detection &  \makecell[ml]{CIFAR-10$^{(\diamond)}$, CIFAR-100$^{(\diamond)}$\\TIM, ImageNet$^{(*)}$} & \xmark & Clusters (Synthetic) \\
            \makecell[tl]{Prior RNN  \citep{shen2020modeling}} & New Concept Extraction & \xmark & \xmark & \makecell[ml]{Concept Learning$^{(\diamond)}$, Snips$^{(\diamond)}$,\\ ATIS$^{(\diamond)}$ (Language)}  \\
            \makecell[tl]{Graph-based Kernel Dirichlet distribution\\estimation (GKDE)
            \citep{zhao2020uncertainty}} & \makecell[ml]{OOD Detection\\ Misclassification Detection} & \xmark & \xmark & \makecell[ml]{Coauthors Physics$^{(\diamond)}$,\\ Amazon Computer$^{(\diamond)}$\\ Amazon Photo$^{(\diamond)}$ (Graph)} \\
            \midrule 
            \makecell[tl]{Evidential Deep Learning\\  \citep{sensoy2018evidential}} & \makecell[ml]{OOD Detection,\\ Adversarial Attack Detection} & \makecell[ml]{MNIST, notMNIST$^{(*)}$,\\CIFAR-10$^{(\dagger)}$} & \xmark & \xmark \\
            \makecell[tl]{Regularized ENN\\ \citet{zhao2019quantifying}} & OOD Detection & \makecell[ml]{CIFAR-10$^{(\dagger)}$} & \xmark & Clusters (Synthetic) \\
            \makecell[tl]{WGAN--ENN\\ \citep{hu2021multidimensional}} & \makecell[ml]{OOD Detection,\\ Adversarial Attack Detection} & \makecell[ml]{MNIST, notMNIST$^{(*)}$,\\CIFAR-10$^{(\dagger)}$} & \xmark & Clusters (Synthetic) \\
            \makecell[tl]{Variational Dirichlet\\ \citep{chen2018variational}} & \makecell[ml]{OOD Detection,\\ Adversarial Attack Detection} & \makecell[ml]{MNIST, CIFAR-10/100,\\ iSUN$^{(*)}$, LSUN$^{(*)}$,\\ SVHN$^{(*)}$, TIM$^{(*)}$} & \xmark & \xmark \\
            \makecell[tl]{Dirichlet Meta-Model \\ \citep{shen2022post}} & \makecell[tl]{OOD Detection\\Misclassification Detection} & \makecell[tl]{MNIST$^{(\diamond,\dagger)}$, CIFAR-10$^{(\diamond,\dagger)}$,\\CIFAR-100$^{(\diamond)}$, Omniglot$^{(*)}$,\\FashionMNIST$^{(*)}$, K-MNIST$^{(*)}$,\\ SVHN$^{(*)}$, LSUN$^{(*)}$,\\TIM$^{(*)}$} & \xmark & \xmark \\
            \makecell[tl]{Belief Matching  \citep{joo2020being}} & \makecell[ml]{OOD Detection, Calibration} & \makecell[ml]{CIFAR-10/100, SVHN$^{(*)}$} & \xmark & \xmark \\ 
            \makecell[tl]{Posterior Networks\\ \citep{charpentier2020posterior}} & \makecell[ml]{OOD Detection,\\ Misclassification Detection,\\Calibration} & \makecell[ml]{MNIST, FashionMNIST$^{(*)}$,\\ K-MNIST$^{(*)}$, CIFAR-10,\\ SVHN$^{(*)}$} & \makecell[ml]{Segment$^{(\dagger)}$,\\ Sensorless Drive$^{(\dagger)}$} & Clusters (Synthetic) \\
            \makecell[tl]{Graph Posterior Networks\\ \citep{stadler2021graph}} & \makecell[ml]{OOD Detection,\\ Misclassification Detection,\\Calibration} & \xmark & \xmark & \makecell[ml]{Amazon Computer$^{(\diamond)}$, Amazon Photo$^{(\diamond)}$\\CoraML$^{(\diamond)}$, CiteSeerCoraML$^{(\diamond)}$,\\ PubMed$^{(\diamond)}$, Coauthors Physics$^{(\diamond)}$,\\ CoauthorsCS$^{(\diamond)}$, OBGN Arxiv$^{(\diamond)}$ (Graph)} \\
            \midrule 
            \makecell[tl]{Deep Evidential Regression \\
            \citep{amini2020deep}} & \makecell[ml]{OOD Detection,\\ Misclassification Detection,\\Adversarial Attack Detection\\Calibration} & \makecell[ml]{NYU Depth v2\\ ApolloScape$^{*}$\\(Depth Estimation)} & \makecell[ml]{UCI Regression\\ Benchmark} & Univariate Regression (Synthetic)\\ 
            \makecell[tl]{Deep Evidential Regression\\with Multi-task Learning\\ \citep{oh2021improving}} & \makecell[ml]{OOD Detection,\\Calibration} & \xmark & \makecell[ml]{Davis, Kiba$^{(\dagger)}$,\\BindingDB, PubChem$^{(*)}$\\(Drug discovery),\\ UCI Regression\\ Benchmark} & Univariate Regression (Synthetic) \\
            \makecell[tl]{ Multivariate Deep Evidential \\ Regression \citet{meinert2021multivariate}} & Qualitative Evaluation & \xmark & \xmark & Multivariate Regression (Synthetic) \\
            \makecell[tl]{Regression Prior Network\\ \citep{malinin2020regression}} & OOD Detection & \makecell[ml]{NYU Depth v2$^{\diamond}$,\\ KITTI$^{\diamond}$\\(Depth Estimation)} & \makecell[ml]{UCI Regression\\ Benchmark} & Univariate Regression (Synthetic) \\
            \makecell[tl]{Natural Posterior Network \\
            \citep{charpentier2021natural}} &  \makecell[ml]{OOD Detection, Calibration} & \makecell[ml]{NYU Depth v2,\\ KITTI$^{*}$, LSUN$^{(*)}$\\(Depth Estimation),\\ MNIST, FashionMNIST$^{(*)}$,\\ K-MNIST$^{(*)}$, CIFAR-10$^{(\dagger)}$,\\ SVHN$^{(*)}$, CelebA$^{(*)}$} & \makecell[ml]{UCI Regression\\ Benchmark$^{(\dagger)}$,\\ Sensorless Drive$^{(\dagger)}$,\\Bike Sharing$^{(\dagger)}$ } & \makecell[ml]{Clusters (Synthetic),\\Univariate Regression (Synthetic))}\\
            \bottomrule
        \end{tabular}%
    }
\end{table}

This section contains a discussion of the used datasets, methods to evaluate the quality of uncertainty evaluation, as well as a direct of available models based on the reported results to determine the most useful choices for practitioners. An overview over the differences between the surveyed works is given in \cref{table:overview-evaluation}.

\paragraph{Datasets} Most models are applied to image classification problems, where popular choices involve the MNIST dataset \citep{lecun1998mnist}, using as OOD datasets Fashion-MNIST \citep{xiao2017fashion}, notMNIST \citep{bulatov2011notmnist} containing English letters, K-MNIST \citep{clanuwat2018deep} with ancient  Japanese Kuzushiji characters, and the Omniglot dataset \citep{lake2015human}, featuring handwritten characters from more than 50 alphabets. Other choices involve different versions of the CIFAR-10 object recognition dataset \citep{lecun1998gradient, krizhevsky2009learning} for training purposes and SVHN \citep{goodfellow2014multi}, iSUN \citep{xiao2010sun}, LSUN \citep{yu2015lsun}, CelebA \citep{liu2015deep}, ImageNet \citep{deng2009imagenet} and TinyImagenet \citep{bastidas2017tiny} for OOD samples. Regression image datasets include for instance the NYU Depth Estimation v2 dataset \citep{silberman2012indoor}, using ApolloScape \citep{huang2018apolloscape} or KITTI \citep{menze2015object} as an OOD dataset. Many authors also illustrate model uncertainty on synthetic data, for instance by simulating clusters of data points using Gaussians \citep{malinin2018predictive,malinin2019reverse, nandy2020towards,zhao2019quantifying,hu2020open,charpentier2020posterior,charpentier2021natural}, spiral data \citep{malinin2020ensemble} or polynomials for regression \citep{amini2020deep, oh2021improving, meinert2021multivariate, malinin2020regression,charpentier2021natural}. Tabular datasets include the Segment dataset, predicting image segments based on pixel features \citep{dua2017uci}, and the sensorless drive dataset \citep{dua2017uci, paschke2013sensorlose}, describing the maintenance state of electric current drives as well as popular regression datasets included in the UCI regression benchmark used by \citet{hernandez2015probabilistic, gal2016dropout}: Boston house prices \citep{harrison1978hedonic}, concrete compression strength \citep{yeh1998modeling}, energy efficiency of buildings \citep{tsanas2012accurate}, forward kinematics of an eight link robot arm \citep{corke1996robotics}, maintenance of naval propulsion systems \citep{coraddu2016machine}, properties of protein tertiary stuctures, wine quality \citep{cortez2009modeling}, and yacht hydrodynamics \citep{gerritsma1981geometry}. Furthermore, \citet{oh2021improving} use a number of drug discovery datasets, such as Davis \citep{davis2011comprehensive}, Kiba \citep{tang2014making}, BindingDB \citep{liu2007bindingdb} and PubChem \citep{kim2019pubchem}. \citet{bilovs2019uncertainty} are the only authors working on asynchronous time even prediction, and supply their own data in the form of processed stack exchange postings, smart home data, and car indicators. \citet{shen2020modeling} provide the sole method on language data, and use three different concept learning datasets, i.e.\@ Concept Learning \citep{jia2017learning}, Snips \citep{coucke2018snips} and ATIS \citep{hemphill1990atis}, which contains new OOD concepts to be learned by design. For graph neural networks, \citet{zhao2020uncertainty} and \citet{stadler2021graph} select data from the co-purchase datasets Amazon Computer, Amazon Photos \citep{mcauley2015image} and the CoraML \citep{mccallum2000automating}, CiteSeer \citep{giles1998citeseer} and PubMed \citep{namata2012query}, Coauthors Physics \citep{shchur2018pitfalls}, CoauthorCS \citep{namata2012query} and OGBN Arxiv \citep{hu2020open} citation datasets. Lastly, \citet{charpentier2021natural} use a single count prediction dataset concerned with predicting the number of bike rentals \citep{fanaee2014event}.\\

\paragraph{Uncertainty Evaluation Methods} There usually are no gold labels for uncertainty estimates, which is why the efficacy of proposed solutions has to be evaluated in a different way. One such way used by almost all the surveyed works is using uncertainty estimates in a proxy OOD detection task: Since the model is underspecified on unseen samples from another distribution, it should be more uncertain. By labelling OOD samples as the positive and ID inputs as the negative class, we can measure the performance of uncertainty estimates using the area under the receiver-operator characteristic (AUROC) or the area under the precision-recall curve (AUPR). We can thereby characterize the usage of data from another dataset as a form of covariate shift, while using left-out classes for testing can be seen as a kind of concept shift \citep{moreno2012unifying}.  Instead of using OOD data, another approach is to use adversarial examples \citep{malinin2019reverse,tsiligkaridis2019information,sensoy2018evidential,hu2021multidimensional,chen2018variational,amini2020deep}, checking if they can be identified through uncertainty. In the case of \citet{shen2020modeling}, OOD detection or new concept extraction is the actual and not a proxy task, and thus can be evaluated using classical metrics such as the $F_1$ score. Another way is misclassification detection: In general, we would desire the model to be more uncertain about inputs it incurs a higher loss on, i.e.\@, what it is more wrong about. For this purpose, some works \citep{malinin2018predictive,zhao2020uncertainty,charpentier2020posterior} measure whether let missclassified inputs be the positive class in another binary proxy classification test, and again measure AUROC and AUPR. Alternatively, \citet{malinin2020ensemble,stadler2021graph,amini2020deep} show or measure the area under the prediction / rejection curve, graphing how task performance varies as predictions on increasingly uncertain inputs is suspended. Lastly, some authors look at a model's calibration \citep{guo2017on}: While this does not allow to judge the quality of uncertainty estimates themselves, quantities like the expected calibration error quantify to what extent the output distribution of a classifier corresponds to the true label distribution, and thus whether aleatoric uncertainty is accurately reflected.\\

%

\clearpage

\section{Fundamental Derivations Appendix}\label{app:fundamental-derivations}

This appendix section walks the reader through generalized versions of recurring theoretical results using Dirichlet distributions in a Machine Learning context, such as their expectation in \cref{app:expectation-dirichlet}, their entropy in \cref{app:entropy-dirichlet} and the Kullback-Leibler divergence between two Dirichlets in \cref{app:infinity-norm-loss}.

\subsection{Expectation of a Dirichlet}\label{app:expectation-dirichlet}

Here, we show results for the quantities $\Expect[]{}{\pi_k}$ and $\Expect[]{}{\log \pi_k}$. For the first, we follow the derivation by \citet{miller2011dirichlet}. Another proof is given by \citet{lin2016dirichlet}.

\begin{align}
    \Expect[]{}{\pi_k} & = \int \cdots \int \pi_k \frac{\Gamma(\alpha_0)}{\prod_{k^\prime=1}^K \Gamma(\alpha_k^\prime)} \prod_{k^\prime=1}^K \pi_{k^\prime}^{\alpha_{k^\prime} - 1} d\pi_1 \ldots d\pi_K \\
    \intertext{Moving $\pi_{k}^{\alpha_{k} - 1}$ out of the product:}
    & = \int \cdots \int \frac{\Gamma(\alpha_0)}{\prod_{k^\prime=1}^K \Gamma(\alpha_{k^\prime})} \pi_k^{\alpha_k - 1 + 1}\prod_{k^\prime \neq k} \pi_{k^\prime}^{\alpha_{k^\prime} - 1}d\pi_1 \ldots d\pi_{K} \\
    \intertext{For the next step, we define a new set of Dirichlet parameters with $\beta_k = \alpha_k + 1$ and $\forall k^\prime \neq k: \beta_{k^\prime} = \alpha_{k^\prime}$. For those new parameters, $\beta_0 = \sum_k \beta_k = 1 + \alpha_0$. So by virtue of the Gamma function's property that $\Gamma(\beta_0) = \Gamma(\alpha_0 + 1) = \alpha_0\Gamma(\alpha_0)$, replacing all terms in the normalization factor yields}
    & = \int \cdots \int \frac{\alpha_k}{\alpha_0}\frac{\Gamma(\beta_0)}{\prod_{k^\prime=1}^K \Gamma(\beta_{k^\prime})} \prod_{k^\prime=1}^K \pi_{k^\prime}^{\beta_{k^\prime} - 1} d\pi_1 \ldots d\pi_K = \frac{\alpha_k}{\alpha_0} 
\end{align}

where in the last step we obtain the final result, since the Dirichlet with new parameters $\beta_k$ must nevertheless integrate to $1$, and the integrals do not regard $\alpha_k$ or $\alpha_0$. For the expectation $\Expect[]{}{\log \pi_k}$, we first rephrase the Dirichlet distribution in terms of the exponential family \citep{kupperman1964probabilities}. The exponential family encompasses many commonly-used distributions, such as the normal, exponential, Beta or Poisson, which all follow the form 

\begin{equation}\label{eq:exp-family}
    p(\bx; \bm{\eta}) = h(\bx)\exp\big(\bm{\eta}^T u(\bx) - A(\bm{\eta})\big)
\end{equation}

with \emph{natural parameters} $\bm{\eta}$, \emph{sufficient statistic} $u(\bx)$, and \emph{log-partition function} $A(\bm{\eta})$. For the Dirichlet distribution, \citet{winn2004variational} provides the sufficient statistic as $u(\bm{\pi}) = [\log \bm{\pi}_1, \ldots, \bm{\pi}_K]^T$ and the log-partition function 

\begin{equation}\label{eq:log-partition-dirichlet}
    A(\bm{\alpha}) = \sum_{k=1}^K \log \Gamma(\alpha_k) - \log \Gamma(\alpha_o) 
\end{equation}

By \citet{introduction2019mao}, we also find that by the moment-generating function that for the sufficient statistic, its expectation can be derived by 

\begin{equation}\label{eq:expected-value-sufficient}
    \Expect[]{}{u(\bx)_k} = \frac{\partial A(\bm{\eta})}{\partial \eta_k}
\end{equation}

Therefore we can evaluate the expected value of $\log \pi_k$ (i.e. the sufficient statistic) by inserting the definition of the log-partition function in \cref{eq:log-partition-dirichlet} into \cref{eq:expected-value-sufficient}:

\begin{equation}\begin{aligned}\label{eq:log-expectation}
    \Expect[]{}{\log \pi_k} = \frac{\partial}{\partial \alpha_k}\sum_{k=1}^K \log \Gamma(\alpha_k) - \log \Gamma(\alpha_0) = \psi(\alpha_k) - \psi(\alpha_0)
\end{aligned}\end{equation}

which corresponds precisely to the definition of the digamma function as $\psi(x) = \frac{d}{d x}\log \Gamma(x)$.

\subsection{Entropy of Dirichlet}\label{app:entropy-dirichlet}

The following derivation is adapted from \citet{lin2016dirichlet}, with the result stated in \citet{charpentier2020posterior} as well.

\begin{align}
    H[\bm{\pi}] & = - \Expect[]{}{\log p(\bm{\pi}|\bm{\alpha})} \\
    & = - \Expect[\bigg]{}{\log\Big( \frac{1}{B(\bm{\alpha})}\prod_{k=1}^K\pi_k^{\alpha_k - 1}\Big)} \\
    & = - \Expect[\bigg]{}{-\log B(\bm{\alpha}) + \sum_{k=1}^K (\alpha_k - 1)\log \pi_k} \\
    & = \log B(\bm{\alpha} ) - \sum_{k=1}^K (\alpha_k - 1)\Expect[]{}{\log \pi_k} \\
    \intertext{Using \cref{eq:log-expectation}:}
    & = \log B(\bm{\alpha} ) - \sum_{k=1}^K (\alpha_k - 1)\big(\psi(\alpha_k) - \psi(\alpha_0)\big) \\
    & = \log B(\bm{\alpha} ) + \sum_{k=1}^K (\alpha_k - 1)\psi(\alpha_0) - \sum_{k=1}^K (\alpha_k - 1)\psi(\alpha_k) \\
    & = \log B(\bm{\alpha}) + (\alpha_0 - K)\psi(\alpha_0) - \sum_{k=1}^K (\alpha_k - 1)\psi(\alpha_k) 
 \end{align}

\subsection{Kullback-Leibler Divergence between two Dirichlets}\label{app:kl-dirichlets}

The following result is presented using an adapted derivation by \citet{lin2016dirichlet} and appears in \citet{chen2018variational} and \citet{joo2020being} as a starting point for their variational objective (see  \cref{app:elbo-dirichlet}). In the following we use $\text{Dir}(\bm{\pi}; \bm{\alpha})$ to denote the optimized distribution, and $\text{Dir}(\bm{\pi}; \bm{\gamma})$ the reference or target distribution.

\begin{align}
    \text{KL}\Big[p(\bm{\pi}|\bm{\alpha})\Big|\Big| p(\bm{\pi}|\bm{\gamma})\Big] = & \Expect[\bigg]{}{\log\frac{p(\bm{\pi}|\bm{\alpha})}{p(\bm{\pi}|\bm{\gamma})}} = \Expect[\bigg]{}{\log p(\bm{\pi}|\bm{\alpha})} - \Expect[\bigg]{}{\log p(\bm{\pi}|\bm{\gamma})} \\
    = & \Expect[\bigg]{}{-\log B(\bm{\alpha}) + \sum_{k=1}^K (\alpha_k -1)\log \pi_k } \nonumber \\
    - & \Expect[\bigg]{}{-\log B(\bm{\gamma}) + \sum_{k=1}^K (\gamma_k -1)\log \pi_k} \\
    \intertext{Distributing and pulling out $B(\bm{\alpha})$ and $B(\bm{\gamma})$ out of the expectation (they don't depend on $\bm{\pi}$):}
    = & - \log\frac{B(\bm{\gamma})}{B(\bm{\alpha})} + \Expect[\bigg]{}{\sum_{k=1}^K (\alpha_k -1)\log \pi_k - (\gamma_k -1)\log \pi_k} \\
    = & - \log\frac{B(\bm{\gamma})}{B(\bm{\alpha})} + \Expect[\bigg]{}{\sum_{k=1}^K (\alpha_k -\gamma_k)\log \pi_k}
    \intertext{Moving the expectation inward and using the identity $\Expect[]{}{\pi_k} = \psi(\alpha_k) - \psi(\alpha_0)$ from \cref{app:expectation-dirichlet}:}
    = & - \log\frac{B(\bm{\gamma})}{B(\bm{\alpha})}+ \sum_{k=1}^K (\alpha_k - \gamma_k)\big(\psi(\alpha_k) - \psi(\alpha_0)\big)
\end{align}

The KL divergence is also used by some works as regularizer by penalizing the distance to a uniform Dirichlet with $\bm{\gamma} = \mathbf{1}$ \citep{sensoy2018evidential}.
In this case, the result above can be derived to be 

\begin{equation}
    \text{KL}\Big[p(\bm{\pi}|\bm{\alpha})\Big|\Big| p(\bm{\pi}|\bm{1})\Big] = \log \frac{\Gamma(K)}{B(\bm{\alpha})} + \sum_{k=1}^K (\alpha_k - 1)\big(\psi(\alpha_k) - \psi(\alpha_0)\big)
\end{equation}

where the $\log \Gamma(K)$ term can also be omitted for optimization purposes, since it does not depend on $\bm{\alpha}$.

\clearpage

\section{Additional Derivations Appendix}\label{app:additional-derivations}

In this appendix we present relevant results in a Machine Learning context, including from some of the surveyed works, featuring as unified notation and annotated derivation steps. These include derivations of expected entropy (\cref{app:expected-entropy}) and mutual information (\cref{app:mutual-information}) as uncertainty metrics for Dirichlet networks. 
Also, we derive a multitude of loss functions, including the $l_\infty$ norm loss of a Dirichlet w.r.t. a one-hot encoded class label in \cref{app:infinity-norm-loss}, the $l_2$ norm loss in \cref{app:il2-norm-loss}, as well as the reverse KL loss by \citet{malinin2019reverse}, the UCE objective \citet{bilovs2019uncertainty, charpentier2020posterior} and ELBO \citet{shen2020modeling, chen2018variational} as training objectives (\cref{app:reverse-kl-loss,app:uce-loss,app:elbo-dirichlet}).

\subsection{Derivation of Expected Entropy}\label{app:expected-entropy}

The following derivation is adapted from \cite{malinin2018predictive} appendix section C.4. In the following, we assume that $\forall k \in \mathbb{K}: \pi_k > 0$:
    
\begin{align}
    & \Expect[\bigg]{p(\bm{\pi}|\bx, \hat{\bm{\theta}})}{H\Big[P(y|\bm{\pi})\Big]} = \int p(\bm{\pi}|\bx, \hat{\bm{\theta}}) \bigg(-\sum_{k=1}^K \pi_k\log \pi_k\bigg) d \bm{\pi} \\
    & = - \sum_{k=1}^K \int p(\bm{\pi}|\bx, \hat{\bm{\theta}})\Big(\pi_k \log \pi_k\Big)d \bm{\pi} \\
    \intertext{Inserting the definition of $p(\bm{\pi}|\bx, \hat{\bm{\theta}}) \approx p(\bm{\pi}|\bx, \mathbb{D})$:}
    & = - \sum_{k=1}^K \Bigg(\frac{\Gamma(\alpha_0)}{\prod_{k^\prime=1}^K \Gamma(\alpha_{k^\prime})}\int \pi_k \log \pi_k \prod_{k^\prime=1}^K\pi_{k^\prime}^{\alpha_{k^\prime} - 1} d\bm{\pi} \Bigg)\\
    \intertext{Singling out the factor $\pi_k$:}
    & = - \sum_{k=1}^K \Bigg(\frac{\Gamma(\alpha_0)}{\Gamma(\alpha_{k})\prod_{k^\prime \neq k} \Gamma(\alpha_{k^\prime})}\pi_k^{\alpha_k-1}\int \pi_k \log \pi_k \prod_{k^\prime \neq k}\pi_{k^\prime}^{\alpha_{k^\prime} - 1} d\bm{\pi} \Bigg)\\
    \intertext{Adjusting the normalizing constant (this is the same trick used in \cref{app:expectation-dirichlet}):}
     & = - \sum_{k=1}^K \Bigg(\frac{\alpha_k}{\alpha_0}\int\frac{\Gamma(\alpha_0+1)}{\Gamma(\alpha_{k}+1)\prod_{k^\prime \neq k} \Gamma(\alpha_{k^\prime})}\pi_k^{\alpha_k-1} \log \pi_k \prod_{k^\prime \neq k}\pi_{k^\prime}^{\alpha_{k^\prime} - 1}  d\bm{\pi} \Bigg)\\
    \intertext{Using the identity $\Expect{}{\log \pi_k} = \psi(\alpha_k) -  \psi(\alpha_0)$ (\cref{eq:log-expectation}). Since the expectation here is w.r.t to a Dirichlet with concentration parameters $\alpha_k + 1$, we obtain}
    & = - \sum_{k=1}^K\frac{\alpha_k}{\alpha_0}\bigg(\psi(\alpha_k+1) -  \psi(\alpha_0+1)\bigg) 
\end{align}
    

\subsection{Derivation of Mutual Information}\label{app:mutual-information}

We start from the expression in \cref{eq:dirichlet-mi}:

\begin{align}
    I\Big[y, \bm{\pi}\Big| \bx, \mathbb{D}\Big] & = H\bigg[\Expect[\Big]{p(\bm{\pi}|\bx, \mathbb{D})}{P(y|\bm{\pi})}\bigg] - \Expect[\bigg]{p(\bm{\pi}|\bx, \mathbb{D})}{H\Big[P(y|\bm{\pi})\Big]} \\
    \intertext{Given that $\Expect[]{}{\pi_k} = \frac{\alpha_k}{\alpha_0}$ (\cref{app:expectation-dirichlet}) and assuming that point estimate $p(\bm{\pi}|\bx, \mathbb{D}) \approx p(\bm{\pi}|\bx, \hat{\bm{\theta}})$ is sufficient \citep{malinin2018predictive}, we can identify the first term as the Shannon entropy $-\sum_{k=1}^K \pi_k \log \pi_k = -\sum_{k=1}^K \frac{\alpha_k}{\alpha_0} \log \frac{\alpha_k}{\alpha_0} $. Furthermore, the second part we already derived in \cref{app:expected-entropy} and thus we obtain:}
    & = -\sum_{k=1}^K \frac{\alpha_k}{\alpha_0}\log \frac{\alpha_k}{\alpha_0} + \sum_{k=1}^K\frac{\alpha_k}{\alpha_0}\bigg(\psi(\alpha_k+1) -  \psi(\alpha_0+1)\bigg) \\
    & = - \sum_{k=1}^K \frac{\alpha_k}{\alpha_0}\bigg(\log \frac{\alpha_k}{\alpha_0} -\psi(\alpha_k+1) + \psi(\alpha_0+1)\bigg)
\end{align}

\subsection{$l_\infty$ Norm Derivation}\label{app:infinity-norm-loss}

In this section we elaborate on the derivation of \citet{tsiligkaridis2019information} deriving a generalized $l_p$ loss, upper-bounding the $l_\infty$ loss. This in turn allows us to easily derive the $l_2$ loss used by \citet{sensoy2018evidential, zhao2020uncertainty}. Here we assume the classification target $y$ is provided in the form of a one-hot encoded label $\by = [\indicator{y = 1}, \ldots, \indicator{y = K}]^T$.

\begin{align}\label{eq:norm-loss}
    \Expect[\big]{p(\bm{\pi}|\bx, \bm{\theta})}{||\by - \bm{\pi}||_\infty} & \le \Expect[\big]{p(\bm{\pi}|\bx, \bm{\theta})}{||\by - \bm{\pi}||_p} \\
    \intertext{Using Jensen's inequality}
    & \le \Big(\Expect[\big]{p(\bm{\pi}|\bx, \bm{\theta})}{||\by - \bm{\pi}||_p^p}\Big)^{1/p} \\
    \intertext{Evaluating the expression with $\forall k\neq y: \mathbf{y}_k = 0$:}
    & = \Big(\Expect[]{}{(1-\pi_y)^p} + \sum_{k \neq y}\Expect[]{}{\pi_{k}^p} \Big)^{1/p}
\end{align}

In order to compute the expression above, we first realize that all components of $\pi$ are distributed according to a Beta distribution $\text{Beta}(\alpha, \beta)$ (since the Dirichlet is a multivariate generalization of the beta distribution) for which the moment-generating function is given as follows:

\begin{equation}
    \Expect[]{}{\pi^p} = \frac{\Gamma(\alpha + p)\Gamma(\beta)\Gamma(\alpha + \beta)}{\Gamma(\alpha + p + \beta)\Gamma(\alpha)\Gamma(\beta)} = \frac{\Gamma(\alpha + p)\Gamma(\alpha + \beta)}{\Gamma(\alpha + p + \beta)\Gamma(\alpha)}
\end{equation}

Given that the first term in \cref{eq:norm-loss} is characterized by $\text{Beta}(\alpha_0-\alpha_y, \alpha_y)$ and the second one by $\text{Beta}(\alpha_k, \alpha_0 - \alpha_k)$, we can evaluate the result in \cref{eq:norm-loss} using the moment generating function:

\begin{align}\label{eq:lp-loss}
   \Expect[\Big]{p(\bm{\pi}|\bx, \bm{\theta})}{||\by - \bm{\pi}||_\infty} & \le \Bigg(\frac{\Gamma(\alpha_0-\alpha_y + p)\Gamma(\alpha_0-\cancel{\alpha_y}+\cancel{ \alpha_y})}{\Gamma(\alpha_0-\cancel{\alpha_y} + p + \cancel{\alpha_y)}\Gamma(\alpha_0-\alpha_y)} + \sum_{k \neq y}\frac{\Gamma(\alpha_k + p)\Gamma(\cancel{\alpha_k} + \alpha_0 - \cancel{\alpha_k})}{\Gamma(\cancel{\alpha_k} + p + \alpha_0 - \cancel{\alpha_k})\Gamma(\alpha_k)} \Bigg)^\frac{1}{p} \\
   & = \Bigg(\frac{\Gamma(\alpha_0-\alpha_y+p)\Gamma(\alpha_0)}{\Gamma(\alpha_0 + p)\Gamma(\alpha_0 - \alpha_y)} + \sum_{k\neq y}\frac{\Gamma(\alpha_k+p)\Gamma(\alpha_0)}{\Gamma(p+\alpha_0)\Gamma(\alpha_k)} \Bigg)^\frac{1}{p} \\
   \intertext{Factoring out common terms:}
    & = \vast(\frac{\Gamma(\alpha_0)}{\Gamma(\alpha_0 + p)}\vast(\frac{\Gamma(\alpha_0-\alpha_y+p)}{\Gamma(\alpha_0 - \alpha_y)} + \sum_{k \neq y}\frac{\Gamma(\alpha_k + p)}{\Gamma(\alpha_k)} \vast)\vast)^\frac{1}{p} \\
    \intertext{Expressing $\alpha_0 - \alpha_k = \sum_{k \neq y}\alpha_k$:}
   & = \bigg(\frac{\Gamma(\alpha_0)}{\Gamma(\alpha_0 + p)}\bigg)^\frac{1}{p}\vast(\frac{\Gamma\Big(\sum_{k \neq y}\alpha_k + p\Big)}{\Gamma\Big(\sum_{k\neq y} \alpha_k\Big)} + \sum_{k \neq y}\frac{\Gamma(\alpha_k + p)}{\Gamma(\alpha_k)} \vast)^\frac{1}{p} 
\end{align}

\subsection{$l_2$ Norm Loss Derivation}\label{app:il2-norm-loss}

Here we present an adapted derivation by \citet{sensoy2018evidential} for the $l_2$-norm loss to train Dirichlet networks. Here we again use a one-hot vector for a label with $\by = [\indicator{y = 1}, \ldots, \indicator{y = K}]^T$.

\begin{align}
    \Expect[\Big]{p(\bm{\pi}|\bx, \bm{\theta})}{||\by - \bm{\pi}||_2^2} & = \Expect[\bigg]{}{\sum_{k=1}^K (\indicator{y = k} - \pi_k)^2} \\
    & = \Expect[\bigg]{}{\sum_{k=1}^K \indicator{y = k}^2 - 2\pi_k\indicator{y = k} + \pi_k^2} \\
    & = \sum_{k=1}^K \indicator{y = k}^2 - 2\Expect[]{}{\pi_k}\indicator{y = k} + \Expect[]{}{\pi_k^2} \\
    \intertext{Using the identity that $\Expect[]{}{\pi_k^2} = \Expect[]{}{\pi_k}^2 + \text{Var}(\pi_k)$:}
    & = \sum_{k=1}^K \indicator{y = k}^2 - 2\Expect[]{}{\pi_k}\indicator{y = k} + \Expect[]{}{\pi_k}^2 + \text{Var}(\pi_k)\\
    & = \sum_{k=1}^K \Big(\indicator{y = k} - \Expect[]{}{\pi_k}\Big)^2 + \text{Var}(\pi_k) \\
    \intertext{Finally, we use the result from \cref{app:expectation-dirichlet} and the result that  $\displaystyle \text{Var}(\pi_k) = \frac{\alpha_k(\alpha_0 - \alpha_k)}{\alpha_0^2(\alpha_0 + 1)}$ (see \citeauthor{lin2016dirichlet}, \citeyear{lin2016dirichlet}):}
    & = \sum_{k=1}^K \Big(\indicator{y = k} -\frac{\alpha_k}{\alpha_0}\Big)^2 + \frac{\alpha_k(\alpha_0 - \alpha_k)}{\alpha_0^2(\alpha_0 + 1)}
\end{align}

\subsection{Derivation of Reverse KL loss}\label{app:reverse-kl-loss}

Here we re-state and annotate the derivation of reverse KL loss by \citet{malinin2019reverse} in more detail, starting from the forward KL loss by \citet{malinin2018predictive}. Note that here, $\hat{\bm{\alpha}}$ contains a dependence on $k$, since \citet{malinin2018predictive} let $\hat{\alpha}_k = \hat{\pi_k}\hat{\alpha}_0$ with $\hat{\alpha}_0$ being a hyperparameter and $\hat{\pi}_k = \indicator{k=y} + (-\indicator{k=y}K + 1)\varepsilon$ and $\varepsilon$ being a small number.

\begin{align}
    & \quad\ \Expect[\bigg]{p(\bx, y)}{\sum_{k=1}^K \indicator{y = k}\text{KL}\Big[p(\bm{\pi}|\hat{\bm{\alpha}})\Big|\Big|p(\bm{\pi}|\bx, \bm{\theta})\Big]} \\ 
    & = \Expect[\bigg]{p(\bx, y)}{\sum_{k=1}^K \indicator{y = k}\int p(\bm{\pi}|\hat{\bm{\alpha}}) \log \frac{p(\bm{\pi}|\hat{\bm{\alpha}})}{p(\bm{\pi}|\bx, \bm{\theta})}d\bm{\pi}} \\ 
    \intertext{Writing the expectation explicitly:}
    & = \int \sum_{k=1}^K p(y=k, \bx)\sum_{k=1}^K \indicator{y = k}\int p(\bm{\pi}|\hat{\bm{\alpha}}) \log \frac{p(\bm{\pi}|\hat{\bm{\alpha}})}{p(\bm{\pi}|\bx, \bm{\theta})}d\bm{\pi} d\bx \\
    & = \int \sum_{k=1}^K p(\bx)P(y=k|\bx)\sum_{k=1}^K \indicator{y = k}\int p(\bm{\pi}|\hat{\bm{\alpha}}) \log \frac{p(\bm{\pi}|\hat{\bm{\alpha}})}{p(\bm{\pi}|\bx, \bm{\theta})}d\bm{\pi} d\bx \\
    & = \Expect[\Bigg]{p(\bx)}{\sum_{k=1}^K P(y=k|\bx)\sum_{k=1}^K\indicator{y = k}\int p(\bm{\pi}|\hat{\bm{\alpha}}) \log \frac{p(\bm{\pi}|\hat{\bm{\alpha}})}{p(\bm{\pi}|\bx, \bm{\theta})}d\bm{\pi}} \\
    \intertext{Adding factor in log, collapsing double sum:}
    & = \Expect[\Bigg]{p(\bx)}{\sum_{k=1}^K P(y=k|\bx)\int p(\bm{\pi}|\hat{\bm{\alpha}}) \log \Bigg( \frac{p(\bm{\pi}|\hat{\bm{\alpha}})\sum_{k=1}^K P(y=k|\bx)}{p(\bm{\pi}|\bx, \bm{\theta})\sum_{k=1}^K P(y=k|\bx)}\Bigg)d\bm{\pi}} \\ 
    \intertext{Reordering, separating constant factor from log:}
    & = \mathbb{E}_{p(\bx)}\Bigg[\int \sum_{k=1}^K P(y=k|\bx)p(\bm{\pi}|\hat{\bm{\alpha}}) \Bigg(\log \bigg( \frac{\sum_{k=1}^K P(y=k|\bx)p(\bm{\pi}|\hat{\bm{\alpha}})}{p(\bm{\pi}|\bx, \bm{\theta})}\bigg )\\
    &  -\underbrace{\log\Big(\sum_{k=1}^K P(y=k|\bx)\Big)}_{=0}\Bigg)d\bm{\pi}\Bigg] \\ 
    & = \Expect[\Bigg]{p(\bx)}{\text{KL}\bigg[\underbrace{\sum_{k=1}^K P(y=k|\bx)p(\bm{\pi}|\hat{\alpha})}_{\text{Mixture of } K \text{ Dirichlets} }\Big|\Big|p(\bm{\pi}|\bx, \bm{\theta}) \bigg]}
\end{align}

 where we can see that this objective actually tries to minimizes the divergence towards a mixture of $K$ Dirichlet distributions. In the case of high data uncertainty, this is claimed incentivize the model to distribute mass around each of the corners of the simplex, instead of the desired behavior shown in  \cref{subfig:simplex-aleatoric}. Therefore, \citet{malinin2019reverse} propose to swap the order of arguments in the KL-divergence, resulting in the following: 

\begin{align}
    & \quad\ \Expect[\bigg]{p(\bx)}{\sum_{k=1}^K P(y=k|\bx)\cdot \text{KL}\Big[p(\bm{\pi}|\bx, \bm{\theta})\Big|\Big|p(\bm{\pi}|\hat{\bm{\alpha}})\Big] } \\
    & = \Expect[\bigg]{p(\bx)}{\sum_{k=1}^K p(y=k|\bx)\cdot \int p(\bm{\pi}|\bx, \bm{\theta})\log \frac{p(\bm{\pi}|\bx, \bm{\theta})}{p(\bm{\pi}|\hat{\bm{\alpha}})}d\bm{\pi}} \\
    \intertext{Reordering:}
    & = \Expect[\bigg]{p(\bx)}{\int p(\bm{\pi}|\bx, \bm{\theta})\sum_{k=1}^K  P(y=k|\bx)\log \frac{p(\bm{\pi}|\bx, \bm{\theta})}{p(\bm{\pi}|\hat{\bm{\alpha}})}d\bm{\pi}} \\
    & = \Expect[\Bigg]{p(\bx)}{\Expect[\bigg]{p(\bm{\pi}|\bx, \bm{\theta})}{\sum_{k=1}^K  P(y=k|\bx)\log p(\bm{\pi}|\bx, \bm{\theta}) - \sum_{k=1}^K  P(y=k|\bx)\log p(\bm{\pi}|\hat{\bm{\alpha}})}} \\
    & = \Expect[\Bigg]{p(\bx)}{\int p(\bm{\pi}|\bx, \bm{\theta})\bigg(\log \bigg(\prod_{k=1}^K p(\bm{\pi}|\bx, \bm{\theta})^{P(y=k|\bx)}\bigg) - \log\bigg(\prod_{k=1}^K  p(\bm{\pi}|\hat{\bm{\alpha}})^{P(y=k|\bx)}\bigg)\bigg)d\bm{\pi}} \\
    & = \mathbb{E}_{p(\bx)}\Bigg[\int p(\bm{\pi}|\bx, \bm{\theta})\bigg(\log \bigg(p(\bm{\pi}|\bx, \bm{\theta})^{\sum_{k=1}^K P(y=k|\bx)}\bigg) 
     - \log\bigg(\prod_{k=1}^K  \Big( \frac{1}{B(\bm{\alpha})}\prod_{k^\prime=1}^K \pi_{k^\prime}^{\alpha_{k^\prime}-1}\Big)^{p(y=k|\bx)}\bigg)\bigg)d\bm{\pi}\bigg] \\
    & = \Expect[\Bigg]{p(\bx)}{\int p(\bm{\pi}|\bx, \bm{\theta})\bigg(\log \big(p(\bm{\pi}|\bx, \bm{\theta})\big) - \log\bigg(\prod_{k=1}^K  \Big( \frac{1}{B(\bm{\alpha})}\prod_{k^\prime=1}^K \pi_{k^\prime}^{\alpha_{k^\prime}-1}\Big)^{P(y=k|\bx)}\bigg)d\bm{\pi}} \\
    & = \Expect[\Bigg]{p(\bx)}{\int p(\bm{\pi}|\bx, \bm{\theta})\bigg(\log \big(p(\bm{\pi}|\bx, \bm{\theta})\big) - \log\bigg( \frac{1}{B(\bm{\alpha})}\prod_{k^\prime=1}^K \pi_{k^\prime}^{\sum_{k=1}^K P(y=k|\bx)\alpha_{k^\prime}-1}\bigg)d\bm{\pi}} \\
    & = \Expect[\Bigg]{p(\bx)}{\text{KL}\Big[p(\bm{\pi}|\bx, \bm{\theta})||p(\bm{\pi}|\bar{\bm{\alpha}})\Big]}\quad\text{where}\quad \bar{\bm{\alpha}}=\sum_{k=1}^K p(y=k|\bx)\alpha_{k^\prime}
\end{align}

Therefore, instead of a mixture of Dirichlet distribution, we obtain a single distribution whose \emph{parameters are a mixture} of the concentrations of each class.\

\subsection{Uncertainty-aware Cross-Entropy Loss}\label{app:uce-loss}

The uncertainty-aware cross-entropy loss in \citet{bilovs2019uncertainty, charpentier2020posterior} has the form

\begin{equation}
    \mathcal{L}_\text{UCE} = \Expect[]{p(\bm{\pi}|\bx, \bm{\theta})}{\log p(y|\bm{\pi})} = \Expect[]{}{\log \pi_y} = \psi(\alpha_y) - \psi(\alpha_0) 
\end{equation}

as $p(y|\bm{\pi})$ is given by the true label in form of a delta distribution, we can apply the result from \cref{app:expectation-dirichlet}.

\subsection{Evidence-Lower Bound For Dirichlet Posterior Estimation}\label{app:elbo-dirichlet}

The evidence lower bound is a well-known objective to optimize the KL-divergence between an approximate proposal and target distribution \citep{jordan1999introduction, kingma2014autoencoding}. We derive it based on \citet{chen2018variational} in the following for the Dirichlet case with a proposal distribution $p(\bm{\pi}|\bx, \bm{\theta})$ to the target distribution $p(\bm{\pi}|y)$. For the first part of the derivation, we omit the dependence on $\bm{\beta}$ for clarity.

\begin{align}
    \text{KL}\big[p(\bm{\pi}|\bx, \bm{\theta})\big|\big|p(\bm{\pi}|y)\big] & = \Expect[\bigg]{p(\bm{\pi}|\bx, \bm{\theta})}{\log\frac{p(\bm{\pi}|\bx, \bm{\theta})}{p(\bm{\pi}|y)}} = \Expect[\bigg]{p(\bm{\pi}|\bx, \bm{\theta})}{\log\frac{p(\bm{\pi}|\bx, \bm{\theta})p(y)}{p(\bm{\pi}, y)}} \\
    \intertext{Factorizing $p(\bm{\pi}, y) = P(y|\bm{\pi})p(\bm{\pi})$, pulling out $p(y)$ as it doesn't depend on $\pi$:}
    & =  \Expect[\bigg]{p(\bm{\pi}|\bx, \bm{\theta})}{\log\frac{p(\bm{\pi}|\bx, \bm{\theta})}{P(y|\bm{\pi})p(\bm{\pi})}} + p(y) \\
    & = \Expect[\bigg]{p(\bm{\pi}|\bx, \bm{\theta})}{\log\frac{p(\bm{\pi}|\bx, \bm{\theta})}{p(\bm{\pi})}} - \Expect[\big]{p(\bm{\pi}|\bx, \bm{\theta})}{\log P(y|\bm{\pi})} + p(y) \\
    & \le  \text{KL}\big[p(\bm{\pi}|\bx, \bm{\theta})\big|\big|p(\bm{\pi})\big] - \Expect[\big]{p(\bm{\pi}|\bx, \bm{\theta})}{\log P(y|\bm{\pi})} 
\end{align}

Now note that the second part of the result is the uncertainty-aware cross-entropy loss from \cref{app:uce-loss} and re-adding the dependence of $p(\pi)$ on $\bm{\gamma}$, we can re-use our result regarding the KL-divergence between two Dirichlets in \cref{app:kl-dirichlets} and thus obtain:

\begin{align}
    \mathcal{L}_\text{ELBO} & = \psi(\beta_y) - \psi(\beta_0) - \log\frac{B(\bm{\beta})}{B(\bm{\gamma})}+ \sum_{k=1}^K (\beta_k - \gamma_k)\big(\psi(\beta_k) - \psi(\beta_0)\big)
\end{align}

which is exactly the solution obtained by both  \citet{chen2018variational} and \citet{joo2020being}.




\clearpage

\section{Overview over Loss Functions Appendix}\label{app:overview-loss-functions}

In \cref{tab:overview-loss,tab:overview-loss2}, we compare the forms of the loss function used by Evidential Deep Learning methods for classification, using the consistent notation from the paper. Most of the presented results can be found in the previous \cref{app:fundamental-derivations} and \cref{app:additional-derivations}. We refer to the original work for details about the objective of \citet{nandy2020towards}.

\newpage

\begin{landscape}
    \begin{table}
        \centering
        \caption{Overview over objectives used by prior networks for classification.}
        \resizebox{1.4\textwidth}{!}{
            \renewcommand{\arraystretch}{2}
            \begin{tabular}{@{}llll@{}}
                \toprule
                Method & Loss function & Regularizer & Comment \\
                \midrule
                \makecell[tl]{Prior networks\\ \citep{malinin2018predictive}}  & $\log\frac{B(\hat{\bm{\alpha}})}{B(\bm{\alpha})}+ \sum_{k=1}^K (\alpha_k - \hat{\alpha}_k)\big(\psi(\alpha_k) - \psi(\alpha_0)\big)$ & $-\log \frac{\Gamma(K)}{B(\bm{\alpha})} + \sum_{k=1}^K (\alpha_k - 1)(\psi(\alpha_k) - \psi(\alpha_0))$  & \makecell[tl]{Target concentration parameters $\hat{\bm{\alpha}}$ are created\\ using a label smoothing approach,\\ i.e. $\hat{\pi}_k = \begin{cases} 1 - (K-1)\varepsilon & \quad \text{if } y = k \\ \varepsilon & \quad \text{if }  y \neq k \end{cases}$.\\ Together with setting $\hat{\alpha}_0$ as a hyperparameter,\\ $\hat{\alpha_k} = \hat{\pi_k}\hat{\alpha_0}$} \\
                \makecell[tl]{Prior networks\\ \citep{malinin2019reverse}} & $\log\frac{B(\hat{\bm{\alpha}})}{B(\bm{\alpha})}+ \sum_{k=1}^K (\alpha_k - \hat{\alpha}_k)\big(\psi(\alpha_k) - \psi(\alpha_0)\big)$ & $\log\frac{B(\bar{\bm{\alpha}})}{B(\bm{\alpha})}+ \sum_{k=1}^K (\alpha_k - \bar{\alpha}_k)\big(\psi(\alpha_k) - \psi(\alpha_0)\big)$ & \makecell[tl]{Similar to above, $\hat{\alpha}_c^{(k)} = \indicator{c = k}\alpha_\text{in} + 1$\\ for in-distribution and $\bar{\alpha}_c^{(k)} = \indicator{c = k}\alpha_\text{out} + 1$\\ where we have hyperparameters set to $\alpha_\text{in} = 0.01$\\ and   $\alpha_\text{out} = 0$. Then finally, $\hat{\bm{\alpha}} = \sum_{k=1}^K p(y=k|\bx)\hat{\alpha}_k$ \\and $\bar{\bm{\alpha}} = \sum_{k=1}^K p(y=k|\bx)\bar{\alpha}_k$.} \\
                \makecell[tl]{Information Robust\\Dirichlet Networks\\ \citep{tsiligkaridis2019information}} & $ \bigg(\frac{\Gamma(\alpha_0)}{\Gamma(\alpha_0 + p)}\bigg)^\frac{1}{p}\vast(\frac{\Gamma\Big(\sum_{k \neq y}\alpha_k + p\Big)}{\Gamma\Big(\sum_{k\neq y} \alpha_k\Big)} + \sum_{k \neq y}\frac{\Gamma(\alpha_k + p)}{\Gamma(\alpha_k)} \vast)^\frac{1}{p}$ & $\frac{1}{2}\sum_{k \neq y}(\alpha_k - 1)^2(\psi^{(1)}(\alpha_k) - \psi^{(1)})(\alpha_0))$ & \makecell[tl]{$\psi^{(1)}$ is the polygamma function defined as\\ $\psi^{(1)}(x) = \frac{d}{d x}\psi(x)$.}\\
                \makecell[tl]{Dirichlet via Function\\Decomposition\\ \citep{bilovs2019uncertainty}} & $\psi(\alpha_y) - \psi(\alpha_0)$ & $\lambda_1\int_0^T \pi_k(\tau)^2 d\tau + \lambda_2 \int_0^T (\nu - \sigma^2(\tau))^2 d\tau$ & \makecell[tl]{Factors $\lambda_1$ and $\lambda_2$ that are treated as hyperparameters\\ that weigh first term pushing the for logit $k$ to zero,\\ while pushing the variance in the first term to $\nu$.} \\ 
                \makecell[tl]{Prior network\\with PAC Reg. \\ \citep{haussmann2019bayesian}} & $-\log \Expect[\bigg]{}{\prod_{k = 1}^K\bigg(\frac{\alpha_k}{\alpha_0}\bigg)^{\indicator{k=y}}}$ & $\sqrt{\frac{ \text{KL}\big[p(\bm{\pi}|\bm{\alpha})\big|\big|p(\bm{\pi}|\mathbf{1})\big] - \log \delta}{N} - 1}$ & \makecell[tl]{The expectation in the loss function is evaluated\\using parameter samples from a weight distribution.\\ $\delta \in [0, 1]$.} \\
                \makecell[tl]{Ensemble Distribution\\ Distillation\\ \citep{malinin2020ensemble}} & \makecell[tl]{$\psi(\alpha_0) - \sum_{k=1}^K \psi(\alpha_k) + \frac{1}{M}\sum_{m=1}^M\sum_{k=1}^K(\alpha_k - 1)$\\ $\log p(y=k|\bx, \bm{\theta}^{(m)})$} & - & \makecell[tl]{The objective uses predictions from a trained ensemble\\ with parameters $\bm{\theta}_1, \ldots, \bm{\theta}_M$.} \\
                \makecell[tl]{Prior networks with\\representation gap\\ \citep{nandy2020towards}} & $-\log \pi_y - \frac{\lambda_\text{in}}{K}\sum_{k=1}^K\sigma(\alpha_k)$ & $-\sum_{k=1} \frac{1}{K} \log \pi_k - \frac{\lambda_\text{out}}{K}\sum_{k=1}^K\sigma(\alpha_k)$ & \makecell[tl]{The main objective is being optimized on in-distribution,\\ the regularizer on out-of-distribution data. $\lambda_\text{in}$ and\\ $\lambda_\text{out}$ weighing terms and $\sigma$ denotes the sigmoid function.} \\
                \makecell[tl]{Prior RNN\\ \citep{shen2020modeling}} & $\sum_{k=1}\indicator{k = y}\log \pi_k$ & $-\log B(\tilde{\bm{\alpha})} + (\hat{\alpha}_0 - K)\psi(\hat{\alpha}_0) - \sum_{k=1}^K (\hat{\alpha}_k - 1)\psi(\hat{\alpha}_k)$ & \makecell[tl]{Here, the entropy regularizer operates on a scaled version of the\\ concentration parameters $\tilde{\bm{\alpha}} = (\mathbf{I}_K - \mathbf{W})\bm{\alpha}$, where $\mathbf{W}$ is learned.} \\
                \makecell[tl]{Graph-based Kernel\\Dirichlet dist. est. (GKDE)\\
                \citep{zhao2020uncertainty}} & $\sum_{k=1}^K \Big(\indicator{y = k} -\frac{\alpha_k}{\alpha_0}\Big)^2 + \frac{\alpha_k(\alpha_0 - \alpha_k)}{\alpha_0^2(\alpha_0 + 1)}$ & $- \log\frac{B(\bm{\alpha})}{B(\hat{\bm{\alpha}})}+ \sum_{k=1}^K (\alpha_k - \hat{\alpha}_k)\big(\psi(\alpha_k) - \psi(\alpha_0)\big)$ & \makecell[tl]{$\hat{\bm{\alpha}}$ here corresponds to a uniform prior including some\\ information about the local graph structure. The authors \\also use an additional knowledge distillation objective,\\ which was omitted here since it doesn't related to the Dirichlet.} \\
                \bottomrule
            \end{tabular}%
        }
        \label{tab:overview-loss}
    \end{table}

\end{landscape}

\begin{landscape}    
    \begin{table}
        \centering
        \caption{Overview over objectives used by  posterior networks for classification.}
        \resizebox{1.4\textwidth}{!}{
            \renewcommand{\arraystretch}{2}
            \begin{tabular}{@{}llll@{}}
                \toprule
                Method & Loss function & Regularizer & Comment \\
                \midrule
                \makecell[tl]{Evidential Deep Learning \\ \citep{sensoy2018evidential}} & $\sum_{k=1}^K \Big(\indicator{y = k} -\frac{\beta_k}{\beta_0}\Big)^2 + \frac{\beta_k(\beta_0 - \beta_k)}{\beta_0^2(\beta_0 + 1)}$ & $-\log \frac{\Gamma(K)}{B(\bm{\beta})} + \sum_{k=1}^K (\beta_k - 1)(\psi(\beta_k) - \psi(\beta_0))$ & \\
                \makecell[tl]{Variational Dirichlet\\ \citep{chen2018variational}} & $\psi(\beta_y) - \psi(\beta_0)$ & $- \log\frac{B(\bm{\beta})}{B(\bm{\gamma})}+ \sum_{k=1}^K (\beta_k - \gamma_k)\big(\psi(\beta_k) - \psi(\beta_0)\big)$ & \\
                \makecell[tl]{Regularized ENN\\ \citet{zhao2019quantifying}} & $\sum_{k=1}^K \Big(\indicator{y = k} -\frac{\beta_k}{\beta_0}\Big)^2 + \frac{\beta_k(\beta_0 - \beta_k)}{\beta_0^2(\beta_0 + 1)}$ & $-\lambda_1 \Expect[\Big]{p_\text{out}(\bx, y)}{\frac{\alpha_y}{\alpha_0}} -\lambda_2 \Expect[\Bigg]{p_\text{confl.}(\bx, y)}{\sum_{k=1}^K \bigg(\frac{\beta_k\sum_{k^\prime \neq k}\beta_{k^\prime}\big( 1 - \frac{|\beta_{k^\prime} - \beta_k|}{\beta_{k^\prime} + \beta_k}\big)}{\sum_{k^\prime \neq k}\beta_{k^\prime}}\bigg)}$ & \makecell[tl]{The first term represents \emph{vacuity}, i.e.\@ the lack of evidence and is\\ optimized using OOD examples. The second term stands for \emph{dissonance},\\ and is computed using points with neighborhoods with different classes\\ from their own. $\lambda_1, \lambda_2$ are hyperparameters.} \\
                \makecell[tl]{WGAN--ENN\\ \citep{hu2021multidimensional}} & $\sum_{k=1}^K \Big(\indicator{y = k} -\frac{\beta_k}{\beta_0}\Big)^2 + \frac{\beta_k(\beta_0 - \beta_k)}{\beta_0^2(\beta_0 + 1)}$ & $-\lambda \Expect[\Big]{p_\text{out}(\bx, y)}{\frac{\alpha_y}{\alpha_0}}$ & \\
                \makecell[tl]{Belief Matching\\ \citep{joo2020being}} & $\psi(\beta_y) - \psi(\beta_0)$ & $- \log\frac{B(\bm{\beta})}{B(\bm{\gamma})}+ \sum_{k=1}^K (\beta_k - \gamma_k)\big(\psi(\beta_k) - \psi(\beta_0)\big)$ & \\ 
                \makecell[tl]{Posterior networks\\ \citep{charpentier2020posterior}} & $\psi(\beta_y) - \psi(\beta_0)$ & $-\log B(\bm{\beta}) + (\beta_0 - K)\psi(\beta_0) - \sum_{k=1}^K (\beta_k - 1)\psi(\beta_k)$ & \\
                \makecell[tl]{Graph Posterior Networks\\ \citep{stadler2021graph}} & $\psi(\beta_y) - \psi(\beta_0)$ & $-\log B(\bm{\beta}) + (\beta_0 - K)\psi(\beta_0) - \sum_{k=1}^K (\beta_k - 1)\psi(\beta_k)$ & \\
                \makecell[tl]{Generative Evidential\\Neural Network\\ \citep{sensoy2020uncertainty}} & $- \sum_{k=1}^K \bigg(\Expect[\big]{p_\text{in}(\bx)}{\log(\sigma(f_{\bm{\theta}}(\bx)))} + \Expect[\big]{p_\text{out}(\bx)}{\log(1 - \sigma(f_{\bm{\theta}}(\bx)))}\bigg)$ & $-\log \frac{\Gamma(K)}{B(\bm{\beta}_{-y})} + \sum_{k\neq y} (\beta_k - 1)(\psi(\beta_k) - \psi(\beta_0))$ & \makecell[tl]{The main loss is a discriminative loss using ID and OOD samples,\\ generated by a VAE. The regularizer is taken over all classes\\ \emph{excluding} the true class $y$ (also indicated by $\bm{\beta}_{-y}$).} \\
                \bottomrule
            \end{tabular}%
            }
        \label{tab:overview-loss2}
    \end{table}
\end{landscape}

\end{document}